\documentclass{article}

\usepackage[english]{babel}

\usepackage[letterpaper,top=2cm,bottom=2cm,left=3cm,right=3cm,marginparwidth=1.75cm]{geometry}

\usepackage{amsmath}
\usepackage{graphicx}
\usepackage{hyperref}
\usepackage{cleveref}
\usepackage{indentfirst}
\usepackage{subcaption}
\usepackage{booktabs}
\usepackage{enumitem}
\usepackage{array}
\usepackage{authblk}
\crefname{equation}{}{}
\crefname{figure}{Fig.}{figures}
\crefname{tabular}{Table}{Tabs.}
\crefname{section}{Section}{Sections}
\crefname{theorem}{Theorem}{Theorems}
\crefname{algorithm}{Algorithm}{Algorithms}

\title{Data-Driven Dimensional Synthesis of Diverse Planar Four-bar Function Generation Mechanisms via Direct Parameterization}

\author[1]{Woon Ryong Kim\thanks{\; equal contribution}}
\author[2]{Jae Heun Jung \protect\footnotemark[1]}
\author[2]{Jeong Un Ha}
\author[2]{Dong Hun Lee}
\author[1]{Jae Kyung Shim\thanks{\; Correspondence: jkshim@korea.ac.kr}}

\affil[1]{Department of Mechanical Engineering, Korea University, Seoul, Republic of Korea}
\affil[2]{Department of Mathemetics, Korea University, Seoul, Republic of Korea}

\begin{document}
\maketitle

\begin{abstract}
\label{Abstract}

Dimensional synthesis of planar four-bar mechanisms is a challenging inverse problem in kinematics, requiring the determination of mechanism dimensions from desired motion specifications. We propose a data-driven framework that bypasses traditional equation-solving and optimization by leveraging supervised learning. Our method combines a synthetic dataset, an LSTM-based neural network for handling sequential precision points, and a Mixture of Experts (MoE) architecture tailored to different linkage types.
Each expert model is trained on type-specific data and guided by a type-specifying layer, enabling both single-type and multi-type synthesis. A novel simulation metric evaluates prediction quality by comparing desired and generated motions. Experiments show our approach produces accurate, defect-free linkages across various configurations. This enables intuitive and efficient mechanism design, even for non-expert users, and opens new possibilities for scalable and flexible synthesis in kinematic design.


\end{abstract}

\section{Introduction}
\label{sec:Introduction}

Planar four-bar mechanisms are widely used in mechanical systems due to their simplicity and versatility.
In designing a mechanism to achieve a desired task, accurately calculating its dimensions—a process known as dimensional synthesis—is essential. 
However, even for four-bar mechanisms, this synthesis presents considerable challenges.
Unlike kinematic analysis, which determines output motion from given dimensions, dimensional synthesis is an inverse problem: given a desired output motion, typically expressed as precision points, the objective is to determine the corresponding mechanism dimensions.

Extensive research has been conducted on dimensional synthesis since Freudenstein~\cite{freudenstein1954analytical} introduced his foundational analytical approach for four-bar mechanisms.
Contemporary studies in this field follow two major approaches: exact synthesis, also known as the precision point approach, which aims to find mechanism dimensions that satisfy desired characteristics exactly only at a finite number of discrete precision points, and the approximate approach, which focuses on obtaining solutions that minimize structural error over the entire range of motion.

In this study, a novel data-driven approach is proposed to solve the dimensional synthesis problem of multi-type four-bar function generation mechanisms, leveraging machine learning to bypass the need to solve complex systems of equations and conduct optimization tasks. 
A supervised learning framework consisting of three key components is proposed: 1) a large synthetic dataset, 2) a deep neural network model, and 3) effective training methods.

To achieve this, datasets are generated through displacement analysis based on established analytical equations.
A customized procedure ensures the effective generation of diverse motion characteristics across all four-bar linkage types, resulting in comprehensive datasets of linkage dimensions and their corresponding motions.
To capture the order-sensitive nature of precision points, the proposed method employs an LSTM architecture, which is well-suited for processing sequential data and does not restrict the number of prescribed precision points.
Furthermore, the Mixture of Experts (MoE) concept is incorporated to construct a unified model comprising multiple LSTM expert models.
Each expert specializes in mapping precision points to linkage dimensions for a specific linkage type.
This structure enables single-type synthesis—solutions for a specific four-bar linkage type—and multi-type synthesis—solutions across all types—through a single training procedure.
It is demonstrated that the proposed approach generates defect-free solutions with minimal error, regardless of the number of precision points or the linkage type.
This capability allows even non-expert designers to explore a wide range of potential mechanisms and select the most suitable solution for their design needs.

In \cref{sec:Kinematics Background}, the fundamental concepts of synthesizing four-bar function generation mechanisms are introduced.
\cref{sec:Data-Driven Method for Dimensional Synthesis} details the approach used to develop a well-trained model, covering dataset generation along with model construction, training, and evaluation.
Sections~\cref{sec:Single-type Synthesis Results} and~\cref{sec:Multi-type Synthesis Results} present the training results, demonstrating the model's performance and providing examples of dimensional synthesis for five and twenty absolute precision points for single-type four-bar mechanisms and the simultaneous generation of multiple-type four-bars, respectively.
\cref{sec:Synthesis Using Relative Precision Points} addresses multi-type four-bar mechanisms with seven and twenty precision points, distinguishing itself from \cref{sec:Multi-type Synthesis Results} by defining the precision points using a relative method.
Finally, \cref{sec:Conclusions} summarizes the results and significance of this study, with further discussions on its potential extensions in \cref{sec:Discussion}.

\subsection*{Our contributions}
\begin{itemize}
    \item We propose pipeline for the data-driven appoach on four-bar dimensional synthesis, consist of dataset construction, neural network model for the synthesis and the novel metric to measure the correctness of the prediction.
    \item For the dataset construction, we propose the novel method for the \textbf{type-specified dataset} consist of four-bar linkages sharing the same types.
    \item We also propose novel layer called \textbf{type specifying layer}, which adjusts the model prediction to have specific type. 
    \item We propose the MoE architecture for the synthesis problem, which is mixture of 16 expert neural networks  trained on each type-specified dataset with corresponding type specifying layer.
    \item To measure the prediction quality, we propose a metric called \textbf{simulation metric}, which measure the consistency between the desired motion and predicted linkage.
    \item In experiments, we show the effectiveness of proposed pipeline for the dimensional synthesis problem that our model can predict the linkage correctly.
\end{itemize}

\section{Related works}
\subsection{synthesis problem}

\subsubsection{Exact synthesis}
Exact synthesis remains the most widely used method for calculating mechanism dimensions by solving a set of loop-closure equations~\cite{sandor1959general} or standard dyad form equations~\cite{erdman1997mechanism} for $n$ prescribed precision points.
Although graphical methods~\cite{norton2003design} are effective for synthesizing mechanisms with a small number of precision points, various analytical methods have been developed for more general dimensional synthesis.
Erdman~\cite{erdman1981three} used dyad loop-closure equations, known as the standard form, for three- and four-precision-point synthesis of either motion generation, path generation with prescribed timing, or function generation.
Freudenstein and Sandor~\cite{freudenstein1959synthesis} focused on five-precision-point synthesis of four-bar path generators.
Suh and Radcliffe~\cite{suh1967synthesis} developed a numerical synthesis method for planar linkages based on displacement matrices.
Huang et al.~\cite{huang2009solving} applied elimination methods for synthesizing four-bar path generation mechanisms with five precision points.
Wampler et al.~\cite{wampler1992complete} combined elimination, multihomogeneous variables, and numerical polynomial continuation to handle nine-precision-point synthesis of four-bar path generators.
Using homotopy continuation methods, Plecnik and McCarthy~\cite{plecnik2014numerical,plecnik2016computational,plecnik2016kinematic} synthesized Watt II, Stephenson II, and Stephenson III six-bar function generation mechanisms for eight, eleven, and nine precision points, respectively.
Almandeel et al.~\cite{almandeel2015function} derived defect-free solutions of slider-crank function generators for four precision points.
Soong and Chang~\cite{soong2011synthesis} proposed a function generation synthesis method of a four-bar with a variable-length driving link, unrestricted by the number of precision points.
Chen and Chiang~\cite{chen1983fourth} developed synthesis techniques for spherical four-bar function generators using relative poles.
Su and McCarthy~\cite{su2007synthesis} synthesized compliant four-bar mechanisms with three prescribed equilibrium positions, and Wang et al.~\cite{wang2021synthesis} presented a numerical procedure for synthesizing a four-bar linkage for tasks involving mixed motion and function generation.

However, exact synthesis methods require significant computational resources when the number of precision points is large, due to the complexity of solving nonlinear equations.
Furthermore, the motion characteristics of mechanisms obtained through these methods may deviate significantly from the desired behavior between precision points.

\subsubsection{Approximate synthesis}

To address the limitations of exact synthesis, numerous studies have explored approximate synthesis methods that impose no restrictions on the number of precision points while minimizing structural error.
Various numerical methods~\cite{plecnik2011five, mclarnan1963synthesis} have been applied to dimensional synthesis, with optimization methods being particularly prevalent, especially those that define appropriate objective functions to minimize structural errors.
Freudenstein~\cite{freudenstein1955approximate} developed an approximate synthesis method for four-bar function generators using a constraint equation that the coupler link length remains constant.
Using least-squares methods, Akhras and Angeles~\cite{akhras1990unconstrained} synthesized planar linkages for the problem of rigid-body guidance, while Suh and Mechlenburg~\cite{suh1973optimal} proposed a technique for the optimal synthesis of planar, spherical, and spatial mechanisms for function generation, path generation, and motion generation.
Shariati and Norouzi~\cite{shariati2011optimal} also applied the least-squares method for the optimal synthesis of four-bar mechanisms to generate a definite mathematical function.
Tinubu and Gupta~\cite{tinubu1984optimal} proposed a method for designing function generation mechanisms without branch defects for an unlimited number of precision points.
Martin et al.~\cite{martin2007mechanism} presented a method to select four-point four-bar motion generators from Burmester solution sets that satisfy Grashof criteria, feasible transmission angles, and a minimum perimeter value.
Li et al.~\cite{li2016novel} introduced an analytical method based on Fourier series components to overcome limitations on the number of precision points for planar four-bar function generation mechanisms.
Copeland et al.~\cite{copeland2021concurrent} presented a continuous approximate synthesis method using specialized algorithms to synthesize planar four-bar function generators, while Javash et al.~\cite{javash2013optimum} and Rao~\cite{rao1979synthesis} explored genetic algorithms and geometric programming, respectively.
Porta et al.~\cite{porta2007box} introduced a box approximation method to compute all possible configurations of planar linkages with an arbitrary number of links.

These approximate approaches offer several advantages over precision point methods.
They provide greater flexibility in selecting the number of precision points, as long as computational resources remain within allowable limits, and they can generate quantitative solutions more efficiently.
However, challenges remain, including dependency on initial values, the risk of convergence to local optima, and the need to reconfigure algorithms or objective functions when changing the target mechanism types (e.g., from double-crank to double-rocker four-bar mechanisms).

Despite the advantages of both exact and approximate approaches, all analytical methods share inherent limitations.
As the number of precision points increases, solving such a system of equations becomes significantly more challenging due to the complex and nonlinear relationship between linkage dimensions and the output motion across $n$ precision points.
This results in increased computation time and uncertainty regarding the existence of a solution, and often leads to failures when no analytical solution exists.
Even when solutions are obtained, they may be singular, imaginary, or otherwise unusable, such as those with defect problems.

\subsection{Machine Learning}
Recent advancements in machine learning offer promising alternatives to traditional analytical methods, particularly for complex or computationally intensive problems, by learning from existing data and providing highly accurate predictions without the need for explicit equations.
Among various techniques, neural networks (NNs)—including deep neural networks (DNNs)~\cite{10.1162/neco.2006.18.7.1527} and convolution neural networks (CNNs)~\cite{lecun1989backpropagation}—have shown strong performance across a wide range of prediction tasks.
For sequential data, recurrent neural networks (RNNs)~\cite{rumelhart1986learning} are often used, though they are limited in learning long-term dependencies due to the gradient vanishing problem~\cite{bengio1994learning}.
To address this, Long Short-Term Memory (LSTM) networks~\cite{lstm} were introduced and have demonstrated excellent performance in regression tasks involving sequential inputs~\cite{shiri2023comprehensive}.

A key advantage of data-driven approaches lies in their ability to leverage large datasets.
In domains where real-world data collection is not feasible, regression tasks often suffer from poor generalization and reduced model performance due to limited data availability~\cite{yue2018synthetic}.
To overcome this, synthetic data has proven effective for training neural networks, including in regression tasks~\cite{yue2018synthetic, sutojo2020investigating, reddy1994using}.
This advantage is particularly relevant in dimensional synthesis, where large-scale datasets containing both precision points and corresponding linkage dimensions can be generated synthetically to support accurate prediction.

\subsubsection{ML for dimensional synthesis}
Various studies have explored the integration of machine learning into dimensional synthesis.
Vasiliu and Yannou~\cite{vasiliu2001dimensional} proposed a case-based approach using a simple feedforward neural network model for synthesizing planar path generation mechanisms, while Deshpande and Purwar~\cite{deshpande2019machine} applied machine learning and computational kinematics to solve defect-free dimensional synthesis using a specialized objective function.
Hoskins and Kramer~\cite{hoskins1993synthesis} and Khan et al.~\cite{khan2015dimensional} used neural networks (NNs) to synthesize four-bar path generators.
Erkaya and Uzmay~\cite{erkaya2009optimization} applied advanced neural networks to the design of a slider-crank mechanism with joint clearance.
More recently, Mo et al.~\cite{mo2020path} used a neural network trained with normalized Fourier descriptors to synthesize high-precision four-bar path generators.
Yim et al.~\cite{yim2021big, yim2023big} presented a big data approach to synthesize planar and spatial mechanism, respectively.
Kapsalyamov et al.~\cite{kapsalyamov2023synthesis} combined computational kinematics with machine learning to synthesize a six-bar mechanism for knee and ankle motion generation.

However, most existing studies focus on synthesizing path and motion generators, with relatively few addressing function generation mechanisms specifically.
Even among those, many are limited to predefined mechanism types, such as slider-crank or four-bar crank-rocker mechanisms.
When narrowing the scope to research that integrates machine learning into the dimensional synthesis of function generators, which is the primary focus of this study, related work is scarce, particularly concerning multi-type four-bar synthesis.

\section{Kinematics Background}
\label{sec:Kinematics Background}

\subsection{Dimensional Synthesis of Four-bar Function Generation Mechanisms}
\label{subsec:Dimensional Synthesis of Four-bar Function Generation Mechanisms}

\begin{figure}[htbp]
\centering
\includegraphics[width=0.8\textwidth]{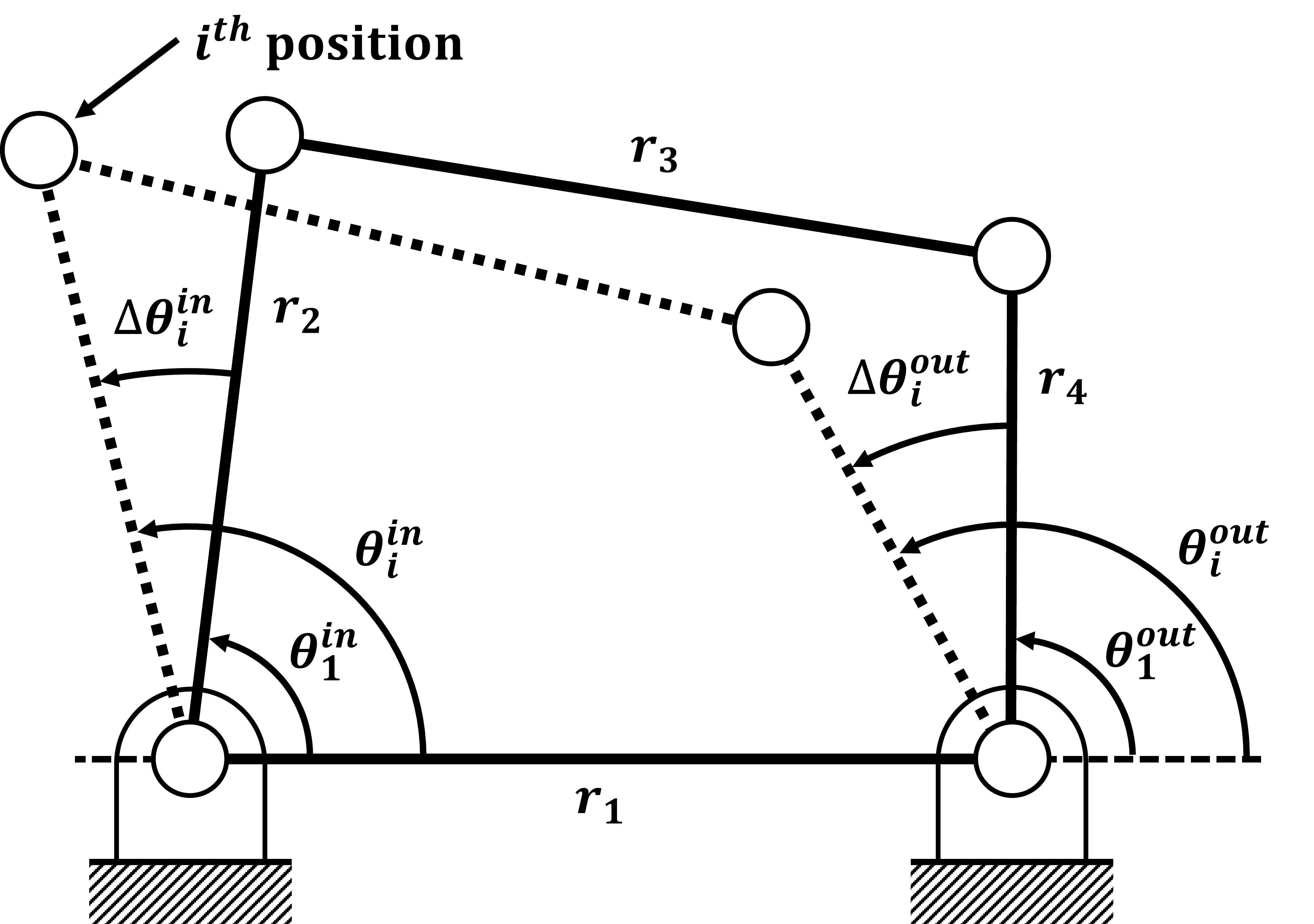}
\caption{Four-bar function generation mechanism}
\label{fig:Four-bar function generation mechanism}
\end{figure}

A planar four-bar linkage, shown in \cref{fig:Four-bar function generation mechanism}, is said to be used as a function generator when the relative rotations between the input and output links, $r_2$ and $r_4$, connected to the ground link, $r_1$, are of interest.
Thus, the objective of dimensional synthesis for a four-bar function generator is to determine the linkage dimensions and initial configuration that satisfy $n$ prescribed precision points, which define the relationship between the input and output link rotations.
The following sections examine the key challenges and considerations in this process.

\subsubsection{Precision Points}
\label{subsubsec:Precision Points}

In four-bar function generator synthesis, a precision point is defined by a pair of rotation angles of the input and the output links.
The $i^{th}$ precision point can be defined in two ways.
If the pair is given by the angular differences between the $i^{th}$ and initial positions of the input and the output links—denoted by $\Delta\theta_i^{in}$ and $\Delta\theta_i^{out}$ in \cref{fig:Four-bar function generation mechanism}—it is referred to as a relative precision point.
In contrast, if represented by the absolute input and output angles at the $i^{th}$ position—denoted by $\theta_i^{in}$ and $\theta_i^{out}$—it is called an absolute precision point.

This study adopts the absolute precision points as the standard representation, as they standardize data format and enhance training efficiency.
Since absolute and relative angles are inherently interchangeable, relative precision points can be considered as a subset of absolute precision points, allowing seamless conversion between the two when necessary.
For completeness, results based on relative precision points are discussed in \cref{sec:Synthesis Using Relative Precision Points}.

In the exact synthesis of four-bar function generators, the maximum number of relative precision points that can be specified is seven, while the maximum number of absolute precision points is five~\cite{erdman1997mechanism, norton2003design}.
Thus, synthesis methods incorporating machine learning to accommodate an unrestricted number of precision points would be highly beneficial.

\subsubsection{Type Classification, Assembly Configuration, and Geometric Inversion}
\label{subsubsec:Type Classification and Assembly Configuration}

As listed in \cref{table:8_types}, four-bar linkages can be classified into eight distinct types~\cite{PICKARD2019237, mccarthy2010geometric, barker1985range} based on the signs of three parameters, $T_1$, $T_2$, and $T_3$.
These are computed from the four defined link lengths $r_1, \ldots, r_4$ shown in \cref{fig:Four-bar function generation mechanism}, where $r_2$ is the input link length and $r_4$ is the output link length, as follows:

\begin{equation}
\label{eq:linkage-types-matrix-form}
    \begin{bmatrix}
        T_1\\T_2\\T_3
    \end{bmatrix}=M_0\vec{r}    
   \; \mbox{ where }\;M_0=\begin{bmatrix}
        1&-1&1&-1\\1&-1&-1&1\\-1&-1&1&1
    \end{bmatrix}  \mbox{ and } \vec{r} = \begin{bmatrix}
        r_1\\r_2\\r_3\\r_4
    \end{bmatrix}
\end{equation}

\begin{table}[htbp]
\centering
    \begin{tabular}{c|c|c|c|c|c}
    \toprule[1.5pt]
    Type no. & Type name & note & $\text{sgn}(T_1)$ &$\text{sgn}(T_2)$&$\text{sgn}(T_3)$ \\ \hline\hline
    1 & Crank-Rocker & - &+&+&+ \\ \hline
    2 & Rocker-Crank & - &+&-&-\\ \hline
    3 & Double-Crank & - &-&-&+\\ \hline
    4 & Double-Rocker & - &-&+&-\\ \hline
    5 & Triple-Rocker & 00 Double-Rocker&-&-&-\\ \hline
    6 & Triple-Rocker & 0$\pi$ Double-Rocker&+&+&-\\ \hline
    7 & Triple-Rocker & $\pi$0 Double-Rocker&+&-&+\\ \hline
    8 & Triple-Rocker & $\pi\pi$ Double-Rocker&-&+&+\\
    \bottomrule[1.5pt]
    \end{tabular}
    \caption{8 types of general four-bar linkage~\cite{PICKARD2019237, mccarthy2010geometric}}
    \label{table:8_types}
\end{table}

Types 1 to 4 in \cref{table:8_types} are Grashof four-bar linkages, each of which can be assembled in two different configurations, referred to as assembly configurations or circuits.
In contrast, Types 5 to 8 are non-Grashof four-bars, each of which has only a single circuit.

\begin{figure}[htbp]
  \centering
  \begin{subfigure}[b]{0.7\textwidth}
    \centering
    \includegraphics[width=\linewidth]{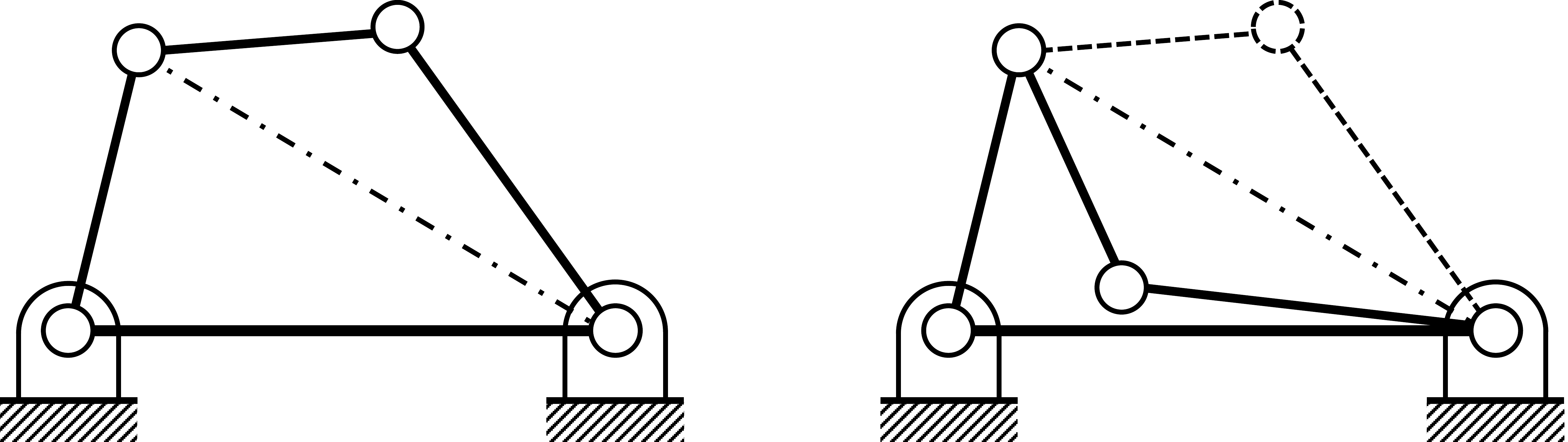}
    \captionsetup{labelformat=parens,labelsep=space}
    \caption{}
    \label{fig:sub1}
  \end{subfigure}
\par\vspace{1em}
  \begin{subfigure}[b]{0.7\textwidth}
    \centering
    \includegraphics[width=\linewidth]{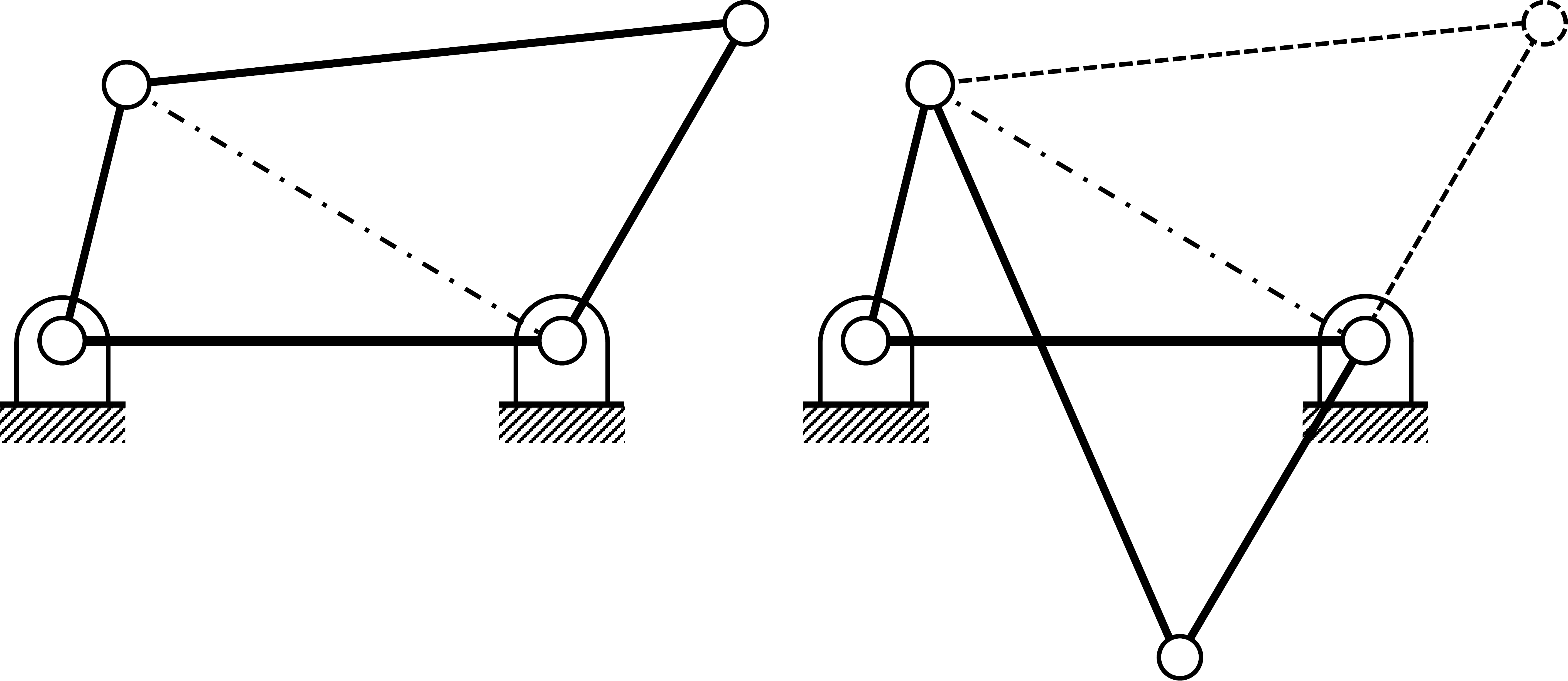}
    \captionsetup{labelformat=parens,labelsep=space}
    \caption{}
    \label{fig:sub2}
  \end{subfigure}
  \caption{Geometric inversions of (a) a rocker-driven four-bar, (b) a crank-driven four-bar}
  \label{fig:Geometric Inversions for each driven type}
\end{figure}

In addition to the circuit, every four-bar linkage has two alternate configurations for a given input position, known as geometric inversions, as illustrated in \cref{fig:Geometric Inversions for each driven type}.
For crank-input four-bars, each geometric inversion corresponds to a distinct circuit, whereas for rocker-input four-bars, both geometric inversions occur within the same circuit.
Since the motion of a four-bar mechanism varies depending on its geometric inversion, this study examines all sixteen possible configurations of four-bar linkages by considering both geometric inversions for each of the eight linkage types.

Note that if any of the parameter $T_j$ for $j \in \{1,2,3\}$ is zero, then the four-bar linkage can have a folding configuration.
Since folding linkages exhibit uncertain configurations, they are excluded from consideration in this study, as their motion cannot be uniquely or reliably defined.

\subsubsection{Circuit, Branch, and Order Defects}
\label{subsubsec:Circuit, Branch, and Order Defects}

Three primary defects are commonly discussed in kinematic synthesis: circuit, branch, and order defects—collectively referred to as CBO defects~\cite{mallik2021kinematic, balli2002defects}.
Circuit and branch defects occur when the mechanism must be disassembled and reassembled to reach all desired positions~\cite{10.1115/1.2919181} while order defects arise when the mechanism reaches the desired positions, but not in the correct order.

These defects can occur in both exact and optimization-based methods, leading to suboptimal or infeasible solutions that fail to achieve the desired motion or effective mechanism design.
Designing defect-free mechanisms remains a core challenge in kinematic synthesis, underscoring the need for reliable methods to ensure desired motions.
In this study, the proposed method yields defect-free solutions, as the dataset is generated using actual moving four-bar mechanisms and constructed through a customized procedure detailed in \cref{sec:Data-Driven Method for Dimensional Synthesis}.

\subsection{Analysis of Four-bar Mechanism}
\label{subsec:Analysis of Four-bar Mechanism}

Since this study proposes a data-driven approach to dimensional synthesis, it is essential to generate datasets that accurately reflect realistic and feasible mechanisms.
Constructing such datasets requires a solid theoretical foundation in kinematic analysis to capture the relationship between linkage parameters and their resulting motions.
Based on this foundation, a database mapping the driver input angle $\theta^{in}$ to the follower output angle $\theta^{out}$ for the four-bar linkage shown in \cref{fig:Design parameters of a four-bar linkage at the ith position} can be generated using the Cartesian form of the loop-closure equation, expressed as follows~\cite{kwak2020kinematic}:

\begin{figure}[htbp]
\centering
\includegraphics[width=0.4\textwidth]{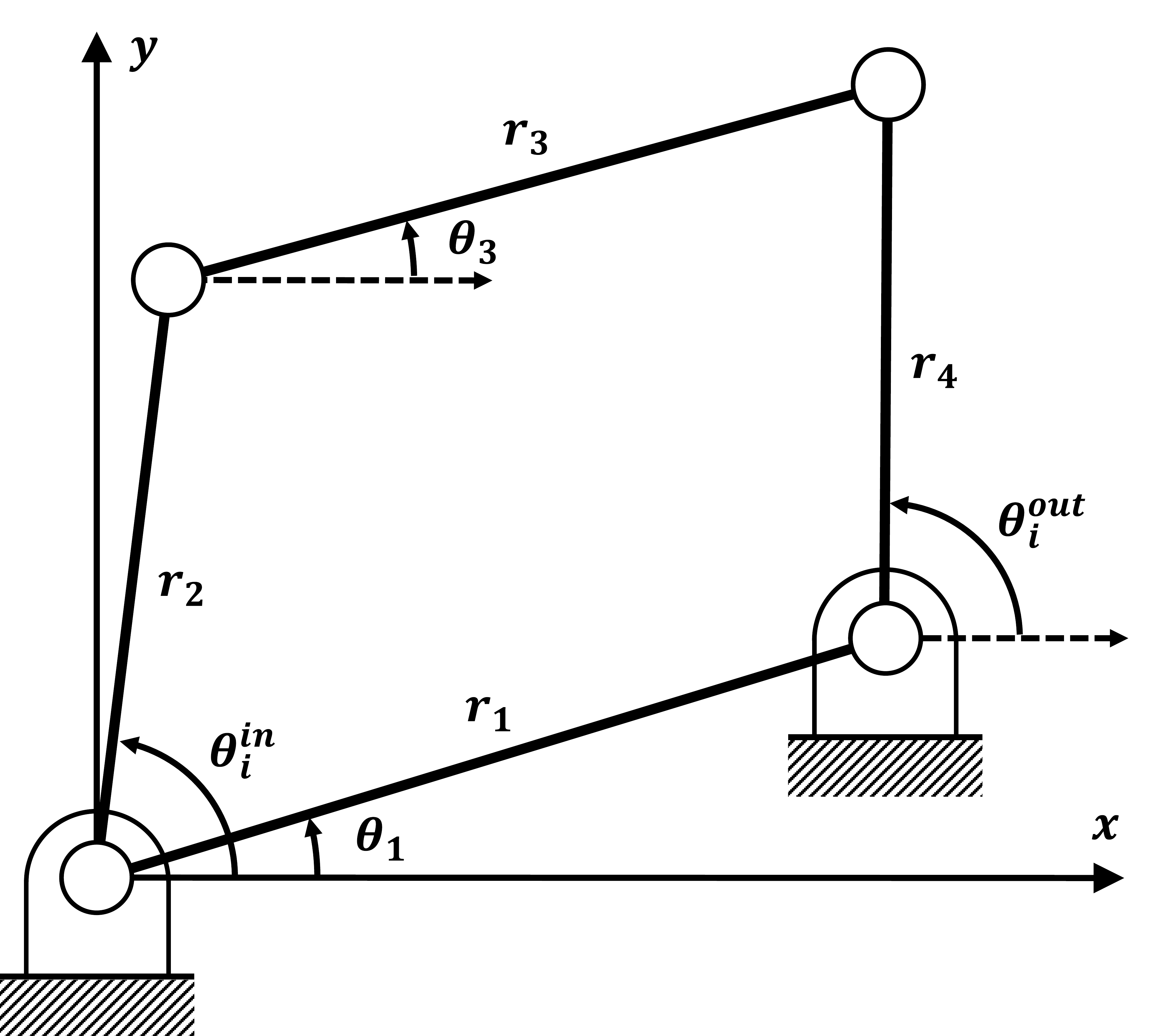}
\caption{Design parameters of a four-bar linkage at the $i^{th}$ position}
\label{fig:Design parameters of a four-bar linkage at the ith position}
\end{figure}

\begin{equation}
\label{eq:4bar-relation}
    \begin{aligned}
        r_2cos\theta^{in}+r_3cos\theta_3 = r_1cos\theta_1+r_4cos\theta^{out}\\
        r_2sin\theta^{in}+r_3sin\theta_3 = r_1sin\theta_1+r_4sin\theta^{out}
    \end{aligned}
\end{equation}

where $\theta_i$ for $i \in \{1,2,3,4\}$, with $\theta_2 = \theta^{in}$ and $\theta_4 = \theta^{out}$, represents the angle of link $r_i$ measured counterclockwise from the positive $x$-axis.
By eliminating $\theta_3$ in \cref{eq:4bar-relation} and applying half-angle identities, the follower output angle $\theta^{out}$ can be determined as follows:

\begin{equation}
\label{eq:theta4_calculation}
    \theta^{out}=2tan^{-1} (\frac{-B\pm\sqrt{B^2-C^2+A^2}}{C-A}), (-\pi\leq \theta^{out}\leq \pi)
\end{equation}

where

\begin{equation}
    \begin{aligned}
        &A=2r_1r_4cos\theta_1-2r_2r_4cos\theta^{in}\\
        &B=2r_1r_4sin\theta_1-2r_2r_4sin\theta^{in}\\
        &C=r_1^2+r_2^2+r_4^2-r_3^2-r_1r_2(cos\theta_1cos\theta^{in}+sin\theta_1sin\theta^{in})
    \end{aligned}
\end{equation}

Note that there are two solutions for $\theta^{out}$ that correspond to two assembly configurations or geometric inversions of the linkage.

Without loss of generality, $\theta_1$ is set to $0^\circ$ in this study.
While the length of $r_1$ could be normalized to 1—since the input-output relationship in a four-bar function generator with rotational input and output is scale-independent~\cite{shariati2011optimal}—this study does not normalize $r_1$ to preserve a broader range of linkage configurations and ensure sufficient diversity in the generated data.

This analytical procedure determines the displacement of the four-bar mechanism for any given set of linkage dimensions, establishing the fundamental relationship between the dimensions and the corresponding input-output rotations in the dataset.

\section{Data-Driven Method for Dimensional Synthesis}
\label{sec:Data-Driven Method for Dimensional Synthesis}

This study proposes a data-driven dimensional synthesis framework for four-bar function generators, leveraging machine learning to infer linkage dimensions from a given set of input-output angular pairs.
In this framework, an LSTM model is employed to learn a mapping from training precision points—defined as a set of $n$ input-output angular pairs $(\bar\theta_i^{in}, \bar\theta_i^{out})$ for $i=1,\ldots,n$—to linkage dimensions $\vec{r} = [r_1, r_2, r_3, r_4]^T$, where each $r_i$ denotes the length of a link in the mechanism.
Both the training precision points and the linkage dimensions are synthetically generated through a custom-designed dataset generation process.

Given a new set of target precision points $(\theta_i^{in}, \theta_i^{out})$ for $i=1,\ldots,n$ specified by a designer, the trained model predicts the corresponding linkage dimensions $\vec{r}_{pred}$, thereby completing the synthesis task.
Since no ground truth solution exists for unseen synthesis problems, the accuracy of the predicted linkage dimensions depends critically on both the quality of the dataset and the effectiveness of the model architecture.

 \subsection{Dataset Generation}
\label{subsec:Dataset Generation}

To represent diverse motion characteristics accurately, this study proposes type-specified datasets, denoted as $\mathcal{D}_{1}^{+}, \ldots, \mathcal{D}_{8}^{+}$ and $\mathcal{D}_{1}^{-}, \ldots, \mathcal{D}_{8}^{-}$, with each dataset corresponding to one of the eight four-bar linkage types described in \cref{subsubsec:Type Classification and Assembly Configuration}.
The subscripts ${+}$ and ${-}$ indicate distinct geometric inversions, determined by sign selection in \cref{eq:theta4_calculation}.
Each dataset entry consists of a pair $(\vec{r}, (\bar\theta_i^{in}, \bar\theta_i^{out}))$, where $\vec{r}$ is a four-dimensional vector representing the ground truth linkage dimensions, and $(\bar\theta_i^{in}, \bar\theta_i^{out})$ is a $2 \times n$ matrix of ground truth training precision points.
The only assumption applied here is that all links are rigid bodies connected by zero-clearance joints, ensuring precise motion without dimensional variation.

The generation process follows three main steps:

\begin{itemize}
\setlength\itemsep{-1.5em}
    \item Step1. Generate linkage dimensions $\vec{r}$ for a specific type-$k$ four-bar mechanism, where $k=1,\ldots,8$, using the $T_j$ parameters as detailed in \cref{subsubsec:Type Specified Dimension Generation}.\\
    \item Step2. Determine the valid input angle range for the driver link and sample $n$ input angles $\bar\theta_i^{in}$ for $i=1,\ldots,n$, as described in \cref{subsubsec:Sampling theta^in}.\\
    \item Step3. Calculate the corresponding output angles $\bar\theta_i^{out}$ for $i = 1, \ldots, n$, as detailed in \cref{subsubsec:Calculating theta^out}. 
\end{itemize}

\begin{figure}[htbp]
\centering
\includegraphics[width=0.7\textwidth]{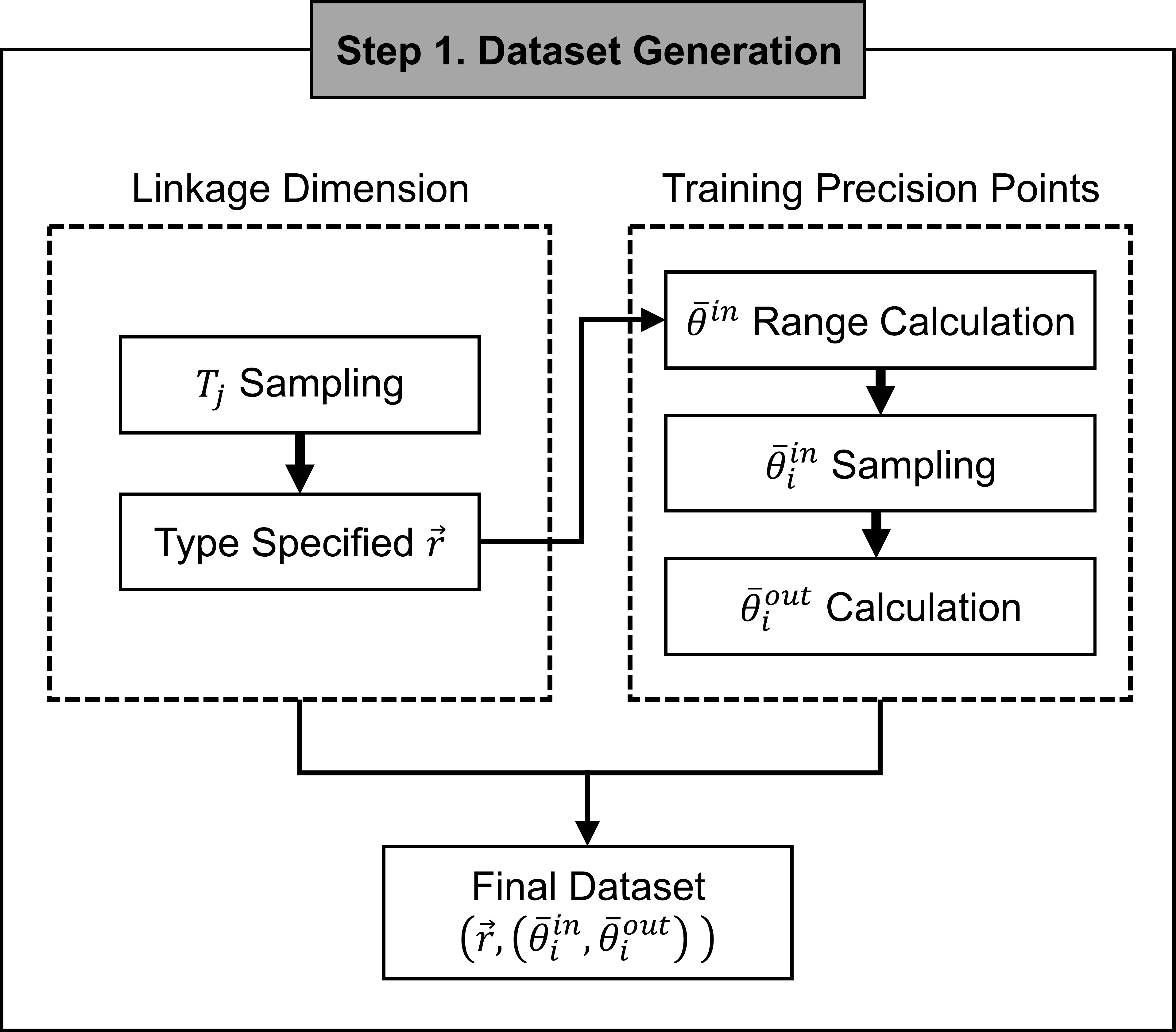}
\caption{Dataset generation process}
\label{fig:Dataset generation process}
\end{figure}

The process of generating a single data sample is illustrated in \cref{fig:Dataset generation process}, and is repeated to construct a large-scale dataset containing a wide variety of linkage types, dimensions, and their corresponding training precision points.
Since the dataset is fully synthetic, its size can theoretically be infinite.
An online training strategy is employed, in which samples are dynamically generated during training to improve generalization and model robustness.

\subsubsection{Type-Specified Dimension Generation}
\label{subsubsec:Type Specified Dimension Generation}

Recall that the type of a four-bar linkage can be determined by the signs of $T_1, T_2,$ and $T_3$, which are given by \cref{eq:linkage-types-matrix-form}, $[T_1, T_2, T_3]^T = M_0\vec{r}$.
By introducing the additional parameter $T_4 = r_1 + r_2 + r_3 + r_4$, $\vec{T} = [T_1, T_2, T_3, T_4]^T$ can be obtained as follows:

\begin{equation}
    \label{eq:T-matrix-form}
    \vec{T}=M\vec{r} \quad \mbox{where}\quad 
    M = \begin{bmatrix}
        1 & -1 & 1 & -1 \\
        1 & -1 & -1 & 1 \\
        -1 & -1 & 1 & 1 \\
        1 & 1 & 1 & 1
    \end{bmatrix}
\end{equation}

Since $M$ is orthogonal with $M^TM = 4I$, the linkage dimensions $\vec{r}$ can be calculated as:

\begin{equation}
\label{eq:r_vec_calculation}
\Vec{r}=\frac{1}{4}M^T\Vec{T} 
\end{equation}

The reason for this formulation, rather than directly randomizing $\vec{r}$, is that even small variations in link lengths can unexpectedly alter the type of a four-bar linkage.
This presents a critical difficulty during training, potentially resulting in significantly different kinematic behavior than expected.
To mitigate this issue, the proposed generation method directly parameterizes $\vec{r}$ by sampling $T_j$ from $[0, \text{sgn}(T_j) \times m]$ for $j \in \{1,2,3,4\}$, generating type-specified linkages and datasets $\mathcal{D}_{k}^{+}$ and $\mathcal{D}_{k}^{-}$ for $k = 1, \ldots, 8$.
Here, $m$ is predefined upper bound on the link length, and the sign values of $T_j$, which determine the linkage type, for $j \in {1,2,3}$ are provided in \cref{table:8_types}.
Note that $\text{sgn}(T_4)$ is always positive by definition.

To ensure $\vec{r}$ forms a valid closed-loop four-bar linkage, the following conditions are checked:

\begin{enumerate}
    \item[(1)] $r_i > 0$ for all $i\in \{1,2,3,4\}$.
    \item[(2)] $2r_i < \sum_{i=1}^{4} r_i$ for all $i\in \{1,2,3,4\}$.
\end{enumerate}

Note that Condition (2) is based on the quadrilateral inequality, which ensures that the length of any side is less than the sum of the remaining three sides.
If the generated $\Vec{r}$ is invalid, it is regenerated until a valid configuration is obtained.

Once $\vec{r}$ is obtained, $n$ training precision points $(\bar\theta_i^{in}, \bar\theta_i^{out})$ for $i=1,\ldots,n$ are constructed.
This is done in two stages: first by sampling appropriate input angles $\bar\theta_i^{in}$ \cref{subsubsec:Sampling theta^in}, and then corresponding output angles $\bar\theta_i^{out}$ are computed \cref{subsubsec:Calculating theta^out}.

\subsubsection{Sampling \texorpdfstring{$\bar\theta_i^{in}$}{theta-in bar}}
\label{subsubsec:Sampling theta^in}

In this step, $n$ input angles $\bar\theta_i^{in}$ are sampled depending on the motion type of the driver input link:

\begin{itemize}
    \item Crank: Sampled from $[-\pi, \pi]$, representing continuous full rotation.
    \item Rocker: Sampled from $[\bar\theta^{in}_{min}, \bar\theta^{in}_{max}] \cup [2\pi + \bar\theta^{in}_{min}, 2\pi + \bar\theta^{in}_{max}]$.
\end{itemize}

\begin{figure}[htbp]
\centering
\begin{subfigure}[b]{0.4\textwidth}
    \centering
    \includegraphics[width=\textwidth]{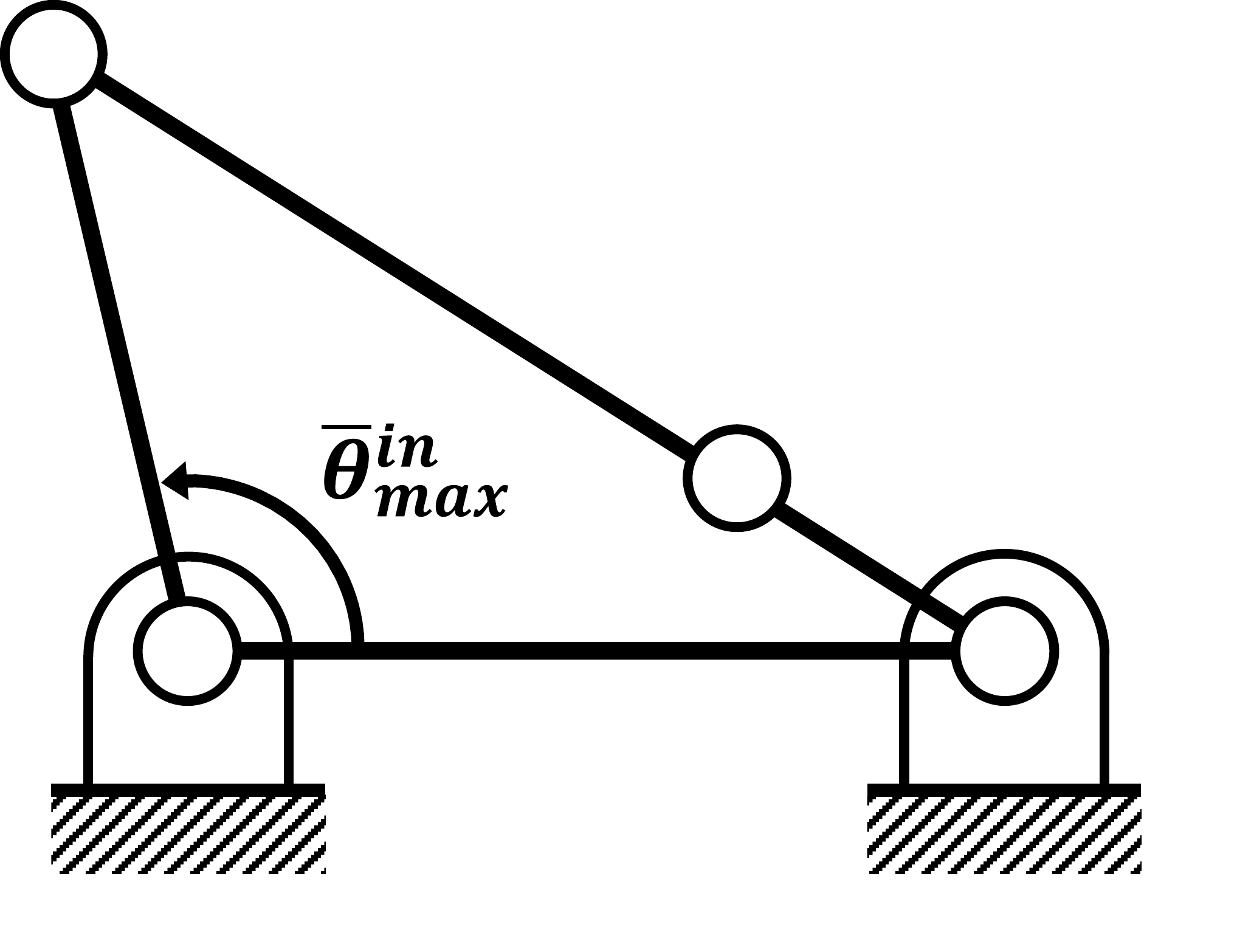}
    \captionsetup{labelformat=parens, labelsep=space}
    \caption*{(a)}
\end{subfigure}
\begin{subfigure}[b]{0.4\textwidth}
    \centering
    \includegraphics[width=\textwidth]{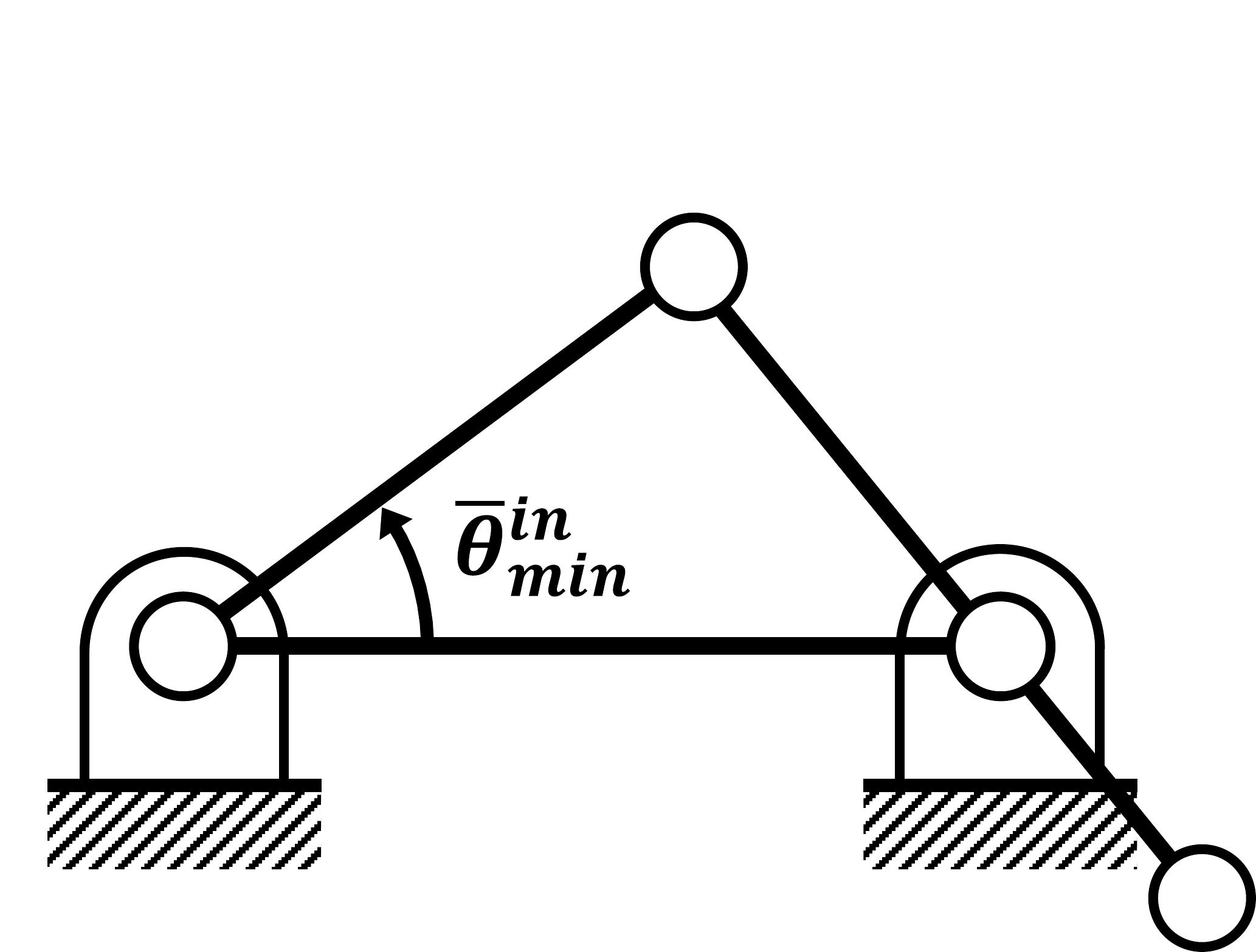}
    \captionsetup{labelformat=parens, labelsep=space}
    \caption*{(b)}
\end{subfigure}
\caption{Dead center positions (DCPs) of four-bar linkage}
\label{fig:Dead center positions (DCPs) of four-bar linkage}
\end{figure}

The rocker range is determined by the configurations at the dead center positions (DCPs) shown in \cref{fig:Dead center positions (DCPs) of four-bar linkage}.
The additional $2\pi$-shifted range is used to resolve ambiguity across DCPs, explained in \cref{subsubsec:Calculating theta^out}.

By defining counterclockwise (CCW) rotation as positive, a complete cycle of a crank driver is a CCW rotation from $-\pi$ to $\pi$.
For a rocker driver, a complete cycle involves a CCW rotation from $\bar\theta^{in}_{min}$ to $\bar\theta^{in}_{max}$, followed by a clockwise (CW) return to $\bar\theta^{in}_{min}$.
Regardless of DCP positions, the DCP that precedes the other in the CCW rotation is defined as $\bar\theta^{in}_{min}$, and the latter as $\bar\theta^{in}_{max}$.
The opposite rotational direction of the driver link is not considered, as a rocker reverses its direction at DCPs, whereas for a crank, the CW rotation is simply the reverse of the CCW input data.

\subsubsection{Calculating \texorpdfstring{$\bar\theta_i^{out}$}{theta-out bar}}
\label{subsubsec:Calculating theta^out}

In the following step, the corresponding output angles $\bar\theta_i^{out}$ for each $\bar\theta_i^{in}$ are computed to complete the training precision points.
Specifically, for crank-input four-bars, such as crank-rocker and double-crank linkages, the output angle $\bar\theta^{out}_i$ is calculated using \cref{eq:theta4_calculation}, with the appropriate sign ($+$ or $-$) determined by the geometric inversion for a given input angle $\bar\theta^{in}_i$ at the $i^{th}$ precision point.
In contrast, for rocker-input four-bars, such as rocker-crank and both Grashof and non-Grashof double-rocker mechanisms that can reach DCPs, determining $\bar\theta_i^{out}$ requires additional considerations.

\begin{figure}[htbp]
\centering
\begin{subfigure}[b]{0.8\textwidth}
    \centering
    \includegraphics[width=\textwidth]{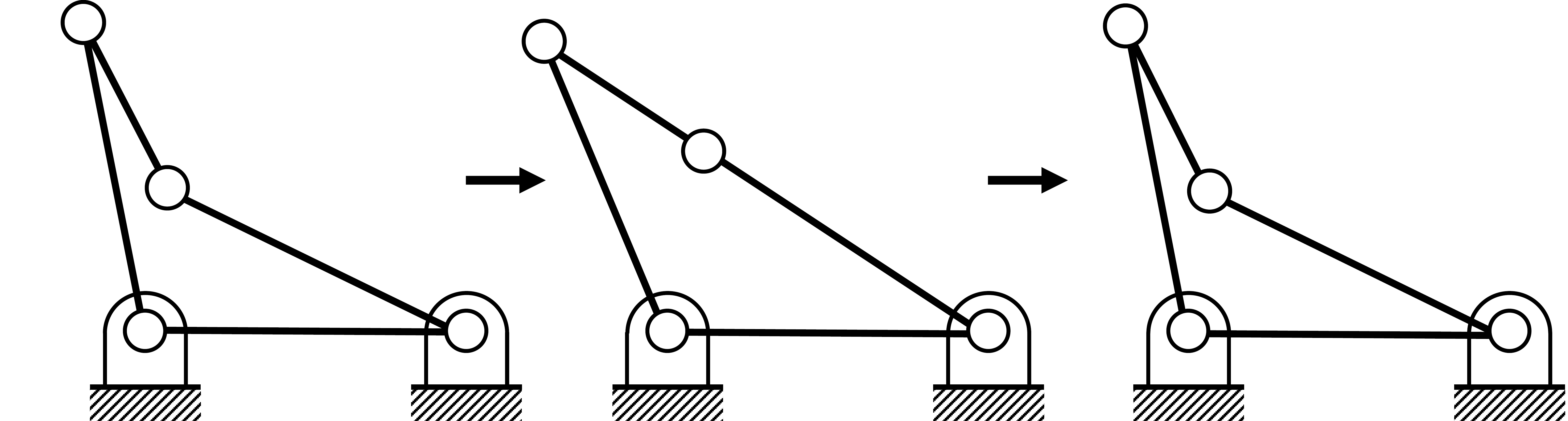}
    \captionsetup{labelformat=parens, labelsep=space}
    \caption*{(a)}
\end{subfigure}
\par\vspace{1em}
\begin{subfigure}[b]{0.8\textwidth}
    \centering
    \includegraphics[width=\textwidth]{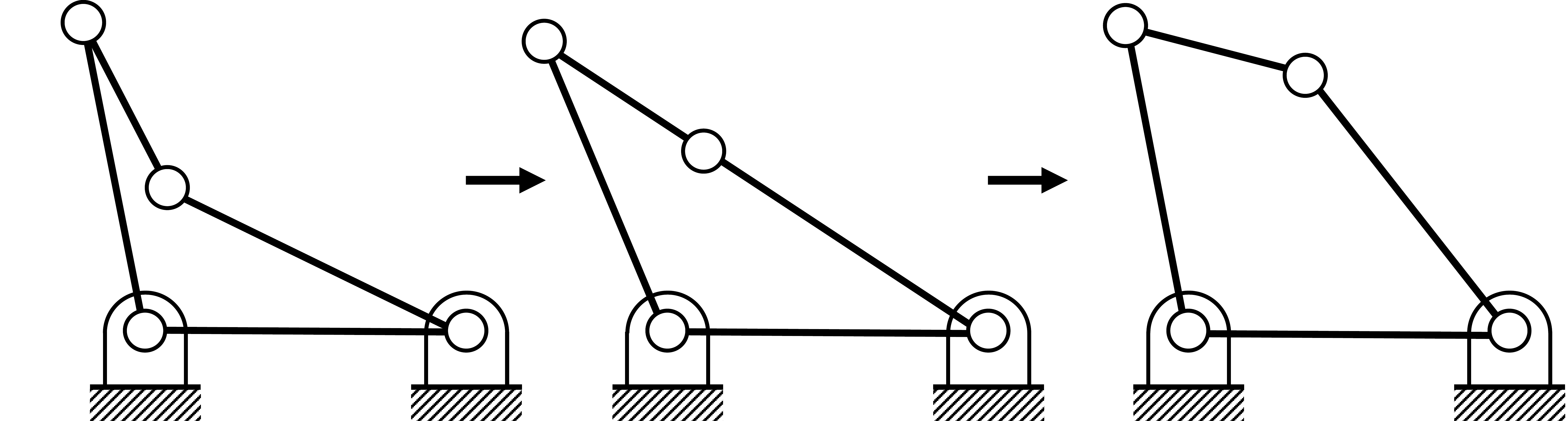}
    \captionsetup{labelformat=parens, labelsep=space}
    \caption*{(b)}
\end{subfigure}

\caption{Two possible motions of a rocker-input four-bar after passing a dead center position (DCP)}
\label{fig:Two possible motions of a rocker-input four-bar after passing a dead center position (DCP)}
\end{figure}

As shown in \cref{fig:Two possible motions of a rocker-input four-bar after passing a dead center position (DCP)}, two possible motions can occur immediately after a rocker-input linkage reaches a DCP, where the radical in \cref{eq:theta4_calculation} becomes zero.
Since both geometric inversions lie within the same circuit, the output link may either return to the previous geometric inversion \cref{fig:Two possible motions of a rocker-input four-bar after passing a dead center position (DCP)}(a) or transition to the opposite geometric inversion \cref{fig:Two possible motions of a rocker-input four-bar after passing a dead center position (DCP)}(b).
In the case of \cref{fig:Two possible motions of a rocker-input four-bar after passing a dead center position (DCP)}(a), the same sign in \cref{eq:theta4_calculation} is used to calculate $\bar\theta_i^{out}$ both before and after the DCP, while in \cref{fig:Two possible motions of a rocker-input four-bar after passing a dead center position (DCP)}(b), the opposite sign must be used after passing the DCP.
Generally, due to the inertia of the output link, the motion depicted in \cref{fig:Two possible motions of a rocker-input four-bar after passing a dead center position (DCP)}(b) is more physically realistic, resembling the motion of a four-bar linkage in a treadle sewing machine.
Therefore, this inertia-based motion is adopted in this study.
However, this approach presents a training challenge, as two different $\bar\theta_i^{out}$ values must be calculated for the same input angle $\bar\theta_i^{in}$ by alternately using both signs in \cref{eq:theta4_calculation}. 

To address this, an additional input angle range of $[2\pi + \bar\theta^{in}_{min}, 2\pi + \bar\theta^{in}_{max}]$ is introduced.
The original range $[\bar\theta^{in}_{min}, \bar\theta^{in}_{max}]$ is used to calculate $\bar\theta_i^{out}$ for CCW input link rotation with the appropriate sign in \cref{eq:theta4_calculation}, while the additional range is used to calculate $\bar\theta_i^{out}$ for CW input link rotation with the opposite sign.
This approach ensures that distinct input values are treated separately by leveraging the periodic nature of angles, thereby resolving issues arising from identical input angles, facilitating efficient training, and preserving the natural motion of the mechanism.

\subsubsection{Final Dataset}
\label{subsubsec:Final Dataset}

Each sample consists of a type-specified linkage dimension $\vec{r}$ and a set of $n$ training precision points $(\bar\theta_i^{in}, \bar\theta_i^{out})$ for $i = 1, \ldots, n$.
The complete dataset is constructed from a diverse collection of such samples, ensuring comprehensive coverage of all valid four-bar linkage types.
Although multi-threaded generation—where multiple CPU cores are used in parallel—is possible, this study employs a single-core CPU setup. This choice avoids unnecessary resource consumption, as single-core performance is sufficient for generating data efficiently. The resulting generation time is 0.07 milliseconds per sample (70 seconds per million samples).

Since this paper presents a data-driven approach involving actual moving mechanisms, all datasets are generated according to specific rules to avoid the CBO defects as follows: a) \textbf{circuit defect}: treating the 8 four-bar linkage types, each with two geometric inversions, as 16 distinct mechanism types; b) \textbf{branch defect}: employing realistic inertia-based motion; c) \textbf{order defect}: treating precision points as sequential entries to the model for training.

\subsection{Model Construction}
\label{subsec:Model Construction}

The synthesis task is formulated as a sequence-to-one regression problem, where a set of training precision points $(\bar\theta_i^{in}, \bar\theta_i^{out})$ is used to predict a single vector $\vec{r}_{pred}$.
To model the sequential characteristics of precision points, an LSTM architecture~\cite{lstm} is employed.
This model processes variable-length sequences, eliminating the need to fix the number of training precision points.
Such flexibility allows for a dynamic number of training precision points, improving generalization across tasks of varying complexity.
Furthermore, by treating training precision points as an ordered sequence, the model captures the inherent sequential structure of the synthesis task, providing robustness against order defects.
While LSTMs can handle sequences of arbitrary length, this study adopts $n = 20$ as a practical upper bound, considering it sufficiently large to represent a wide range of synthesis cases while maintaining computational efficiency.

Given the significant variation in the behavior of the resulting motion across different linkage types, a Mixture of Experts (MoE) framework is adopted.
This structure consists of 16 type-specific LSTM expert models, each exclusively trained on the dataset corresponding to a predefined linkage type.
Each expert model benefits from the LSTM architecture’s flexibility, supporting training with arbitrary numbers of precision points $n$ and enabling unified training across different synthesis cases without requiring separate models for specific settings (e.g., five-point synthesis).
Further architectural details are provided in \cref{subsubsec:Mixture of Experts}.

\subsubsection{Training}
\label{subsubsec:Training}

\begin{figure}[htbp]
\centering
\includegraphics[width=0.7\textwidth]{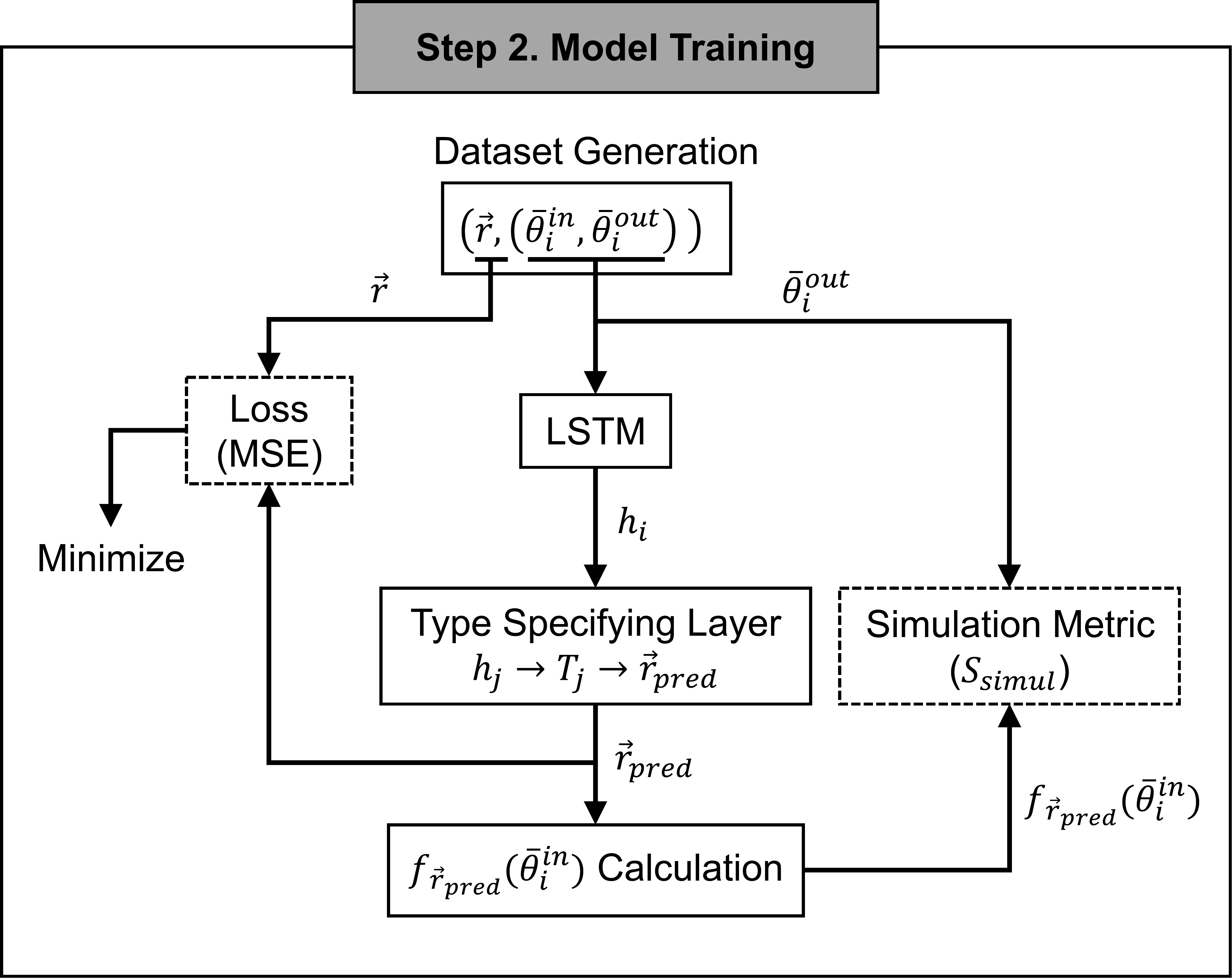}
\caption{Model training process}
\label{fig:Model training process}
\end{figure}

Each expert LSTM model is trained using the procedure illustrated in \cref{fig:Model training process}, based on a large-scale dataset generated through the method in \cref{subsec:Dataset Generation}.

Given a set of training precision points, each expert model predicts a vector $\vec{r}_{pred}$ that satisfies the constraints of its designated linkage type.
This is achieved by a type-specifying layer that transforms the raw LSTM outputs $h_1, h_2, h_3, h_4$ into $T_j$ as follows:

\begin{equation}
    T_j = Softplus(h_j)\times sgn(T_j) \mbox{ for } j=1,2,3,4
\end{equation}

Here, $\text{Softplus}$\cite{softplus} is defined as $\ln(1 + e^x)$, ensuring non-negative values, while $\text{sgn}(T_j)$, provided in \cref{table:8_types},  enforces the appropriate sign based on the target linkage type, as detailed in \cref{subsubsec:Type Specified Dimension Generation}.
The final predicted linkage vector $\vec{r}_{pred}$ is then computed using \cref{eq:r_vec_calculation}, guaranteeing type validity.

All expert models are trained independently using the Mean Squared Error (MSE) loss between $\vec{r}_{pred}$ and the ground truth $\vec{r}$ provided in the dataset, as defined in \cref{eq:MSE}.

\begin{equation}
\label{eq:MSE}
\mathcal{L}_{\mathrm{MSE}} = \frac{1}{4} \sum_{i=1}^{4} \left( \vec{r}_i - (\vec{r}_{pred})_i \right)^2
\end{equation}

\subsubsection{Metrics}
\label{subsubsec:Metrics}

To assess model performance, two evaluation metrics are considered: the cosine similarity metric and the simulation metric.

Cosine similarity is a widely used metric in deep learning that measures the angular similarity between two vectors.
It is invariant to scale, making it suitable for assessing how closely the predicted $\vec{r}_{pred}$ aligns with the ground truth $\vec{r}$ in function generation tasks, where the input-output relationship remains unchanged under uniform scaling of linkage dimensions.
The cosine similarity metric is defined as follows:

\begin{equation}
    S_{cos}(\Vec{r}, \Vec{r}_{pred}) = \frac{\Vec{r} \cdot \Vec{r}_{pred}}{\lVert \Vec{r} \rVert \lVert \Vec{r}_{pred} \rVert}
\end{equation}

where $\lVert\vec{r}\rVert$ denotes the Euclidean norm of $\vec{r}$.

In contrast, the simulation metric $S_{simul}$, measures how closely the predicted $\vec{r}_{pred}$ replicates the target angular relationships defined by the precision points.
It is defined as follows:

\begin{equation}
    S_{simul}(\Vec{r}_{pred},(\bar\theta_i^{in},\bar\theta_i^{out})) = 1-\frac{1}{n}\sum_{i=1}^{n} cos(\bar\theta_i^{out}-f_{\Vec{r}_{pred}}(\bar\theta_i^{in})).
\end{equation}

Here, $f_{\vec{r}_{pred}}$ denotes the angular transformation realized by the mechanism configured with $\vec{r}_{pred}$, mapping each input angle to a corresponding output angle.
This metric evaluates the alignment between the simulated output angles $f_{\vec{r}_{pred}}(\bar\theta_i^{in})$ and the ground truth output angles $\bar\theta_i^{out}$, which are part of the training precision point.
A key advantage of $S_{simul}$ is that it does not require access to $\vec{r}$, making it well-suited for monitoring during training.
Moreover, its cosine-based formulation naturally accounts for angular periodicity, preventing error divergence caused by a 360-degree phase difference.
Note that $S_{simul}$ is subtracted from 1 to align with standard minimization objectives during optimization, so that values closer to 0 indicate better performance.

Due to its effectiveness and independence from ground truth linkage dimensions, $S_{simul}$ is adopted as the evaluation criterion in this study.

\subsubsection{Mixture of Experts}
\label{subsubsec:Mixture of Experts}

\begin{figure}[htbp]
\centering
\includegraphics[width=0.7\textwidth]{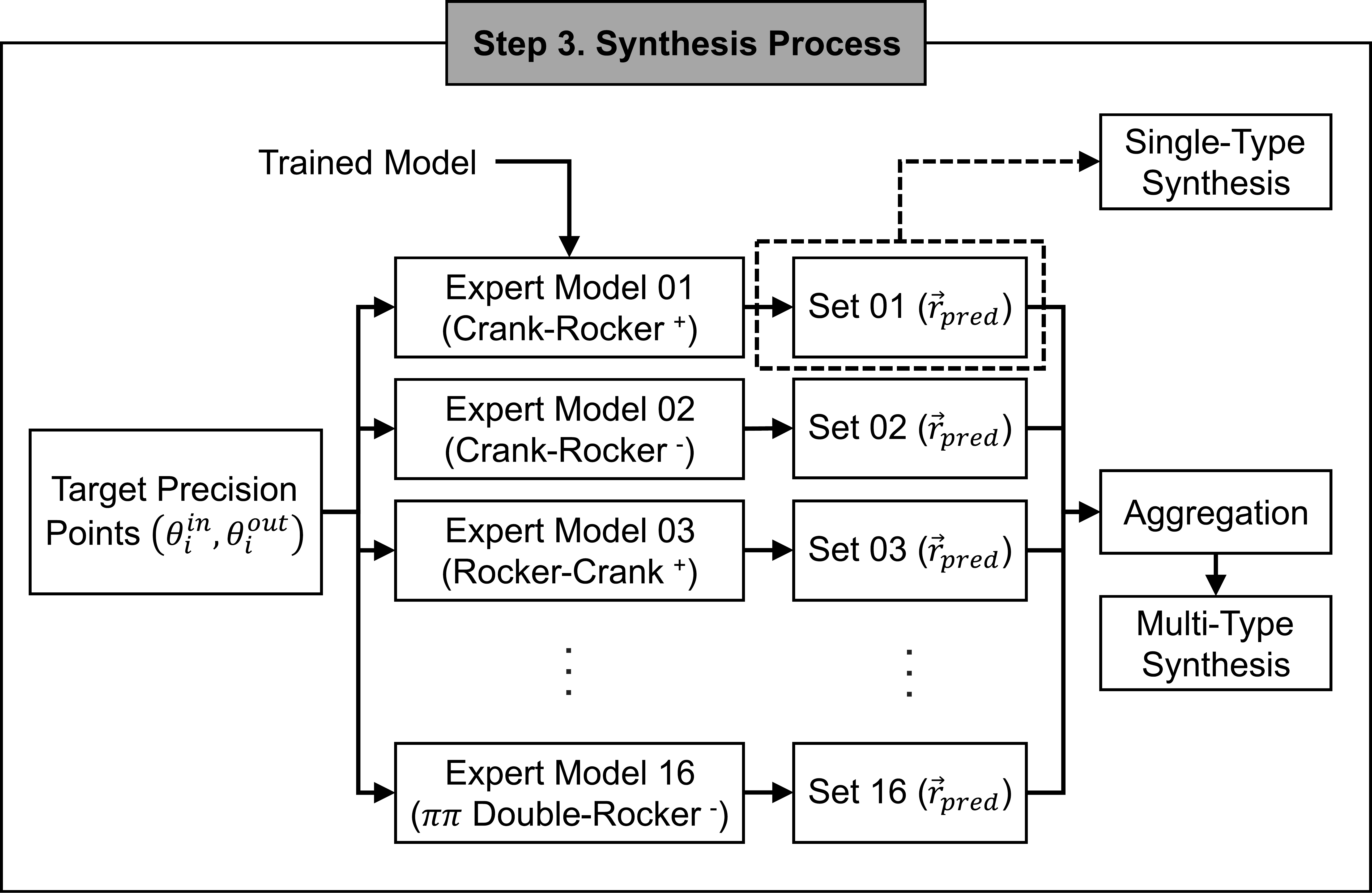}
\caption{Mixture of Experts architecture}
\label{fig:Mixture of Experts architecture}
\end{figure}

The overall architecture of the MoE framework is illustrated in \cref{fig:Mixture of Experts architecture}, while the individual training process of the expert models is shown in \cref{fig:Model training process}.

At the prediction phase, all 16 expert models generate type-specified predictions $\vec{r}_{pred}$.
These predictions are then ranked using the simulation metric $S_{simul}$, and top-scoring predictions are selected. \
This structure supports both type-specified (single-type) synthesis—enabling multiple solutions within a single linkage type—and broad multi-type exploration across all types, thereby increasing both reliability and diversity in the resulting linkage configurations.

\subsubsection{Implementation Details}
\label{subsubsec:Implementation Details}

A unidirectional LSTM architecture with seven layers is employed, each with a dropout probability of 0.3, followed by a linear prediction head.
The model is trained using the Adam optimizer with a weight decay of $2 \times 10^{-3}$ and an initial learning rate of $10^{-4}$.
A multi-step learning rate scheduler is applied, with milestones at epoch 200 and a decay factor $\gamma = 0.8$ to improve training stability and convergence.
All expert models are independently trained for 2000 epochs on an NVIDIA RTX 4090 GPU, with each epoch processing 1024 data samples.
Under this configuration, a complete training session involving all 16 expert models typically takes 10 to 15 hours, depending on system load and training dynamics.
Once training is complete, each expert model can instantly generate mechanism dimensions for its corresponding linkage type, given a set of target precision points, with negligible computation time for aggregating results across all experts.

\section{Single-type Synthesis Using Absolute Precision Points}
\label{sec:Single-type Synthesis Results}

The proposed framework for four-bar function generator synthesis is implemented as a Mixture of Experts (MoE), consisting of 16 expert models, each specialized for a specific linkage type.
Accordingly, the performance of each individual expert directly contributes to the overall synthesis capability of the system.
To evaluate the performance of individual experts, this section presents results for single-type synthesis, using only the expert model designated for a specific linkage type, with given target precision points.

The simulation metric $S_{simul}$, introduced in \cref{subsubsec:Metrics}, is summarized for all expert models in \cref{table:Simulation Error Table}.
Each expert is evaluated on a type-specified dataset generated using the same procedure described in \cref{subsec:Dataset Generation}, except that the ground truth linkage dimensions $\vec{r}$ are withheld during evaluation.
The two distinct geometric inversions of the same linkage type are denoted as $+$ and $-$ in the table, respectively.

An important consideration is that, in practical design scenarios, designers may not know whether a specific linkage type is physically capable of realizing the desired motion.
If synthesis is attempted with an infeasible type—where no realizable mechanism exists for the target precision points—the simulation metric may appear abnormally large, while still generating linkage dimensions that approximate the target motion as closely as possible within the constraints of that type.
These outcomes, however, do not indicate model failure but rather reflect the physical limitation that no valid solution exists within the specified linkage type.

\begin{table}[htbp]
\centering
\begin{tabular}{|l||c|l||c|}
\hline
\multicolumn{1}{|c||}{\textbf{Mechanism Type}} & \multicolumn{1}{c|}{\textbf{Simulation Metric}} & \multicolumn{1}{|c||}{\textbf{Mechanism Type}} & \multicolumn{1}{c|}{\textbf{Simulation Metric}} \\
\hline
Crank Rocker$^{+}$ & 0.0001 & Triple Rocker 00$^{+}$ & 0.0016\\
Crank Rocker$^{-}$ & 0.0001 & Triple Rocker 00$^{-}$ & 0.0015\\
Rocker Crank$^{+}$ & 0.0186 & Triple Rocker 0$\pi$$^{+}$ & 0.0027\\
Rocker Crank$^{-}$ & 0.0179 & Triple Rocker 0$\pi$$^{-}$ & 0.0028\\
Double Crank$^{+}$ & 0.0001 & Triple Rocker $\pi$0$^{+}$ & 0.0026\\
Double Crank$^{-}$ & 0.0001 & Triple Rocker $\pi$0$^{-}$ & 0.0025\\
Double Rocker$^{+}$ & 0.0001 & Triple Rocker $\pi\pi$$^{+}$ & 0.0021\\
Double Rocker$^{-}$ & 0.0001 & Triple Rocker $\pi\pi$$^{-}$ & 0.0012\\
\hline
\end{tabular}
\caption{Simulation metric for 16 types of four-bar mechanisms using absolute precision points}
\label{table:Simulation Error Table}
\end{table}

To further demonstrate the effectiveness of the proposed approach, \cref{subsec:Five Precision Point Case_single} and \cref{subsec:20 Precision Point Case_single} provide synthesis examples using five and twenty absolute precision points, respectively.
As discussed in \cref{subsubsec:Precision Points}, the maximum number of absolute precision points that can be specified is five.
For this reason, five-precision-point synthesis problems are selected as examples, and twenty-point examples are also provided to demonstrate the applicability of the proposed method to cases with a large number of precision points, without increasing computation time.
Each example includes the target precision points $(\theta^{in}$, $\theta^{out})$, the predicted linkage dimensions $\vec{r}_{pred}$, the corresponding output angles $\theta^{out}_{pred}$, and the absolute error at each point (in degrees).
All angles are assumed to lie within the range
$[-\pi, \pi].$ 
However, in the displacement plots that visualize the angular relationship between the generated mechanism and the target precision points, some angles are adjusted from those in the tables to ensure continuity, accounting for the periodicity of rotational angles.

\begin{equation}
\label{eq:absolute error}
\mbox{Absolute Error} \; (\theta^{out},\theta_{pred}^{out}) = \left| \theta_{pred}^{out} - \theta^{out} \right|
\end{equation}

\subsection{Five Precision Point Synthesis}
\label{subsec:Five Precision Point Case_single}

\begin{figure}[htbp]
\centering
\begin{subfigure}[b]{0.45\textwidth}
    \centering
    \includegraphics[width=\textwidth]{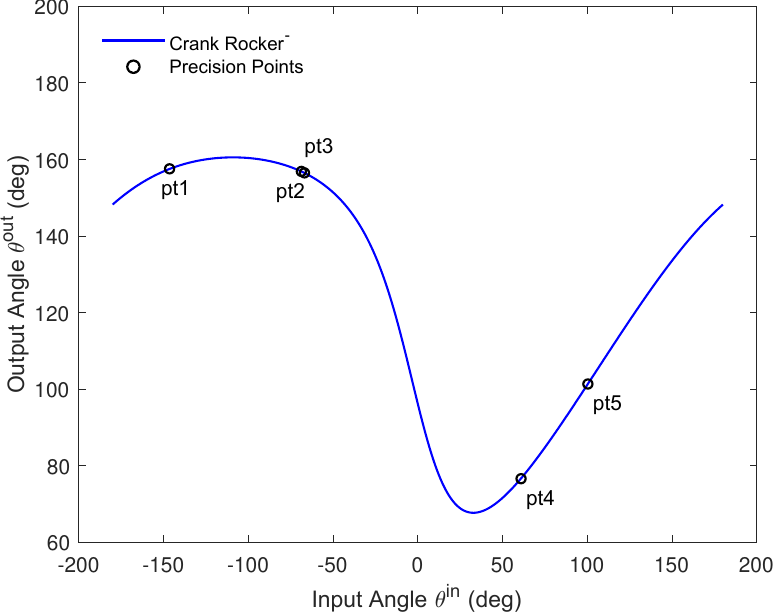}
    \caption*{(a)}
\end{subfigure}
\hfill
\begin{subfigure}[b]{0.45\textwidth}
    \centering
    \includegraphics[width=\textwidth]{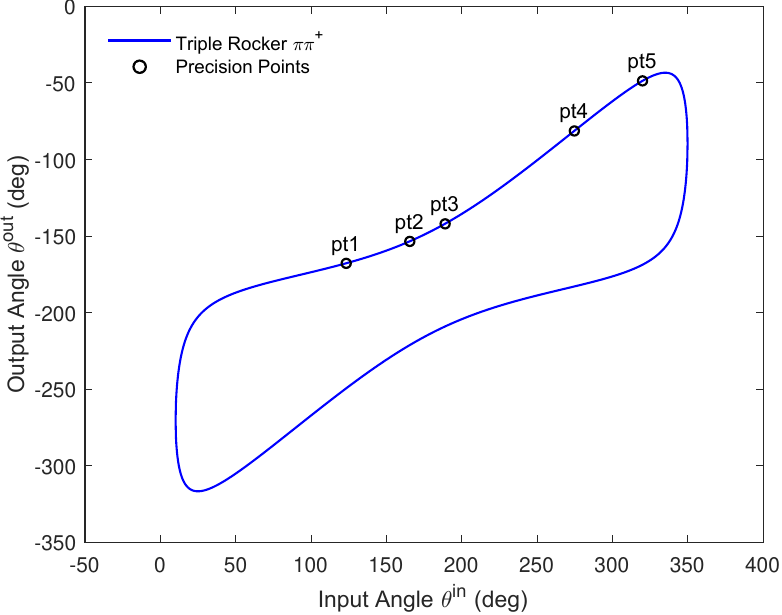}
    \caption*{(b)}
\end{subfigure}
\caption{Displacement plots for five-precision-point synthesis: (a) Crank-Rocker$^{-}$, (b) Triple Rocker $\pi\pi$$^{+}$}
\label{figure:Single_example_five}
\end{figure}

For the five arbitrary precision points listed in \cref{table:Single_example_five_1}, the proposed method synthesized a crank-rocker$^{-}$ mechanism with dimensions $\Vec{r}_{pred} = [2.39072, 2.43180, 2.77589, 3.20339]^T$.
The displacement plot in \cref{figure:Single_example_five}(a) demonstrates that the function generator passes through all five precision points, with a maximum absolute error of $0.03233^{\circ}$.
Since the precision points are specified in absolute angles, the initial orientations of the input link $r_2$ and output link $r_4$ correspond to $\theta^{in}$ and $\theta^{out}_{pred}$ of the $1^{st}$ precision point shown in \cref{table:Single_example_five_1}.
Another example, described in \cref{table:Single_example_five_2}, generated a triple rocker $\pi\pi$$^{+}$ mechanism with dimensions $\Vec{r}_{pred} = [3.02590, 1.94110, 2.94831, 2.86078]^T$.
Its displacement diagram, shown in \cref{figure:Single_example_five}(b), required shifting the $3^{rd}, 4^{th}, 5^{th}$ precision points by $360$ degrees to ensure continuity in plotting.
\begin{table}[htbp]
\centering
\resizebox{\textwidth}{!}{%
\begin{tabular}{|c|c c c c c|}
\hline
\textbf{Precision points} & \textbf{1} & \textbf{2} & \textbf{3} & \textbf{4} & \textbf{5}\\
\hline
\textbf{$\theta^{in}$ (deg)} & -146.41770 & -68.67784 & -66.86595 & 60.90528 & 100.32907\\
\hline
\textbf{$\theta^{out}$ (deg)} & 157.56737 & 156.89046 & 156.50664 & 76.67048 & 101.35825\\
\hline
\textbf{$\theta_{pred}^{out}$ (deg)} & 157.53504 & 156.87263 & 156.48918 & 76.69750 & 101.38479\\
\hline
\textbf{Absolute error (deg)} & 0.03233 & 0.01783 & 0.01746 & 0.02702 & 0.02654\\
\hline
\end{tabular}%
}
\caption{Precision points, output angles, and error for crank rocker$^{-}$ mechanism}
\label{table:Single_example_five_1}
\end{table}

\begin{table}[htbp]
\centering
\resizebox{\textwidth}{!}{%
\begin{tabular}{|c|c c c c c|}
\hline
\textbf{Precision points} & \textbf{1} & \textbf{2} & \textbf{3} & \textbf{4} & \textbf{5}\\
\hline
\textbf{$\theta^{in}$ (deg)} & 123.33112 & 165.54825 & -171.01668 & -85.26373 & -40.01676\\
\hline
\textbf{$\theta^{out}$ (deg)} & -167.81787 & -153.49615 & -142.01949 & -81.34822 & -48.64816\\
\hline
\textbf{$\theta_{pred}^{out}$ (deg)} & -167.75768 & -153.40075 & -141.91932 & -81.29774 & -48.62504\\
\hline
\textbf{Absolute error (deg)} & 0.06019 & 0.09540 & 0.10017 & 0.05048 & 0.02312\\
\hline
\end{tabular}%
}
\caption{Precision points, output angles, and error for triple rocker $\pi\pi^{+}$ mechanism}
\label{table:Single_example_five_2}
\end{table}

\subsection{Twenty Precision Point Synthesis}
\label{subsec:20 Precision Point Case_single}

\begin{figure}[htbp]
\centering
\begin{subfigure}[b]{0.45\textwidth}
    \centering
    \includegraphics[width=\textwidth]{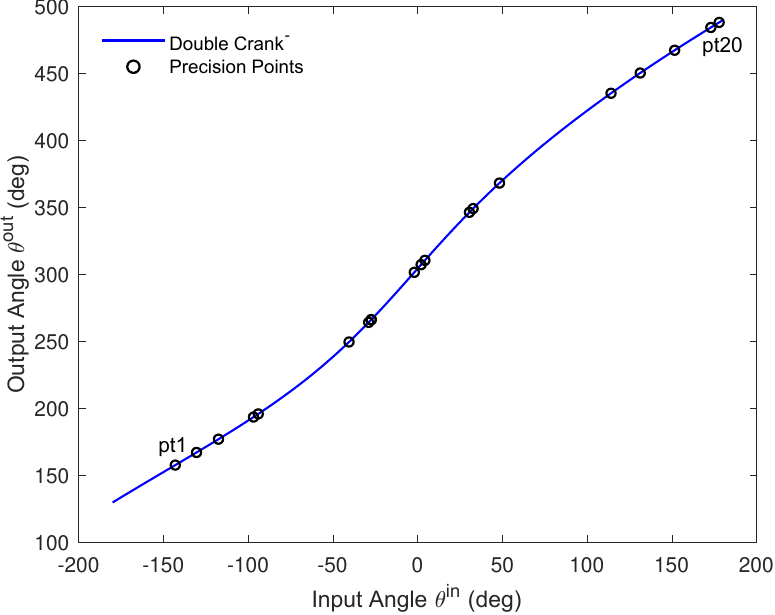}
    \caption*{(a)}
\end{subfigure}
\hfill
\begin{subfigure}[b]{0.45\textwidth}
    \centering
    \includegraphics[width=\textwidth]{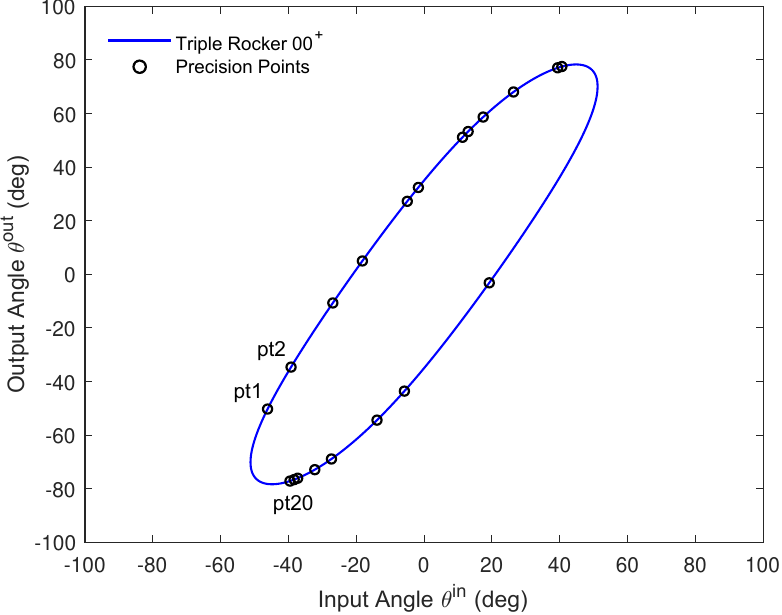}
    \caption*{(b)}
\end{subfigure}
\caption{Displacement plots for twenty-precision-point synthesis: (a) Double Crank$^{-}$ (b) Triple Rocker 00$^{+}$}
\label{figure:Single_example_20}
\end{figure}

For the twenty precision points given in \cref{table:Single_example_20_1}, a double crank$^{-}$ mechanism with dimensions $\vec{r}_{pred} = [0.78543, 2.62035, 2.98265, 3.60855]^T$ is determined, and its displacement plot is shown in \cref{figure:Single_example_20}(a), with a maximum absolute error of $0.01268^{\circ}$.
For the values in \cref{table:Single_example_20_2}, a triple rocker 00$^{+}$ mechanism with dimensions $\Vec{r}_{pred} = [1.63001, 4.72686, 1.83387, 2.08299]^T$ is determined and its displacement plot is shown in \cref{figure:Single_example_20}(b), with a maximum absolute error of $0.02080^{\circ}$.
Note that, for clarity, not all precision points are labeled in \cref{figure:Single_example_20}, however, their order is properly aligned.
These two examples demonstrate the performance of the proposed model in handling a larger number of precision points with high accuracy.

\begin{table}[htbp]
\centering
\resizebox{\textwidth}{!}{%
\begin{tabular}{|>{\centering\arraybackslash}p{3.7cm}|c c c c c c c c c c|}
\hline
\textbf{Precision points} & \textbf{1} & \textbf{2} & \textbf{3} & \textbf{4} & \textbf{5} & \textbf{6} & \textbf{7} & \textbf{8} & \textbf{9} & \textbf{10}\\
\hline
\textbf{$\theta^{in}$ (deg)} & -143.06040 & -130.52878 & -117.57024 & -96.90057 & -94.12099 & -40.58180 & -28.98180 & -27.52273 & -27.43652 & -2.05514\\
\hline
\textbf{$\theta^{out}$ (deg)} & 157.64510 & 167.06403 & 176.96822 & -166.49928 & -164.17302 & -110.47327 & -95.83572 & -93.91568 & -93.80171 & -58.46772\\
\hline
\textbf{$\theta_{pred}^{out}$ (deg)} & 157.64320 & 167.06175 & 176.96540 & -166.50326 & -164.17720 & -110.48244 & -95.84619 & -93.92630 & -93.81234 & -58.48027\\
\hline
\textbf{Absolute error (deg)} & 0.00190 & 0.00228 & 0.00282 & 0.00398 & 0.00418 & 0.00917 & 0.01047 & 0.01062 & 0.01063 & 0.01255\\
\hline
\end{tabular}
}

\resizebox{\textwidth}{!}{%
\begin{tabular}{|>{\centering\arraybackslash}p{3.7cm}|c c c c c c c c c c|}
\hline
\textbf{Precision points} & \textbf{11} & \textbf{12} & \textbf{13} & \textbf{14} & \textbf{15} & \textbf{16} & \textbf{17} & \textbf{18} & \textbf{19} & \textbf{20}\\
\hline
\textbf{$\theta^{in}$ (deg)} & 1.94462 & 4.19549 & 30.45585 & 32.60674 & 48.18012 & 113.96455 & 131.18367 & 151.50618 & 172.85342 & 177.76192\\
\hline
\textbf{$\theta^{out}$ (deg)} & -52.75654 & -49.55105 & -13.69008 & -10.93130 & 8.13623 & 75.14704 & 90.27618 & 107.25335 & 124.24271 & 128.04906\\
\hline
\textbf{$\theta_{pred}^{out}$ (deg)} & -52.76919 & -49.56373 & -13.70164 & -10.94268 & 8.12636 & 75.14307 & 90.27320 & 107.25121 & 124.24107 & 128.04747\\
\hline
\textbf{Absolute error (deg)} & 0.01265 & 0.01268 & 0.01156 & 0.01138 & 0.00987 & 0.00397 & 0.00298 & 0.00214 & 0.00164 & 0.00159 \\
\hline
\end{tabular}
}
\caption{Precision points, output angles, and error for double crank$^{-}$ mechanism}
\label{table:Single_example_20_1}
\end{table}

\begin{table}[htbp]
\centering
\resizebox{\textwidth}{!}{%
\begin{tabular}{|c|c c c c c c c c c c|}
\hline
\textbf{Precision points} & \textbf{1} & \textbf{2} & \textbf{3} & \textbf{4} & \textbf{5} & \textbf{6} & \textbf{7} & \textbf{8} & \textbf{9} & \textbf{10}\\
\hline
\textbf{$\theta^{in}$ (deg)} & -46.15300 & -39.22699 & -26.90189 & -18.18792 & -4.95008 & -1.67552 & 11.34628 & 13.00010 & 17.43324 & 26.38015\\
\hline
\textbf{$\theta^{out}$ (deg)} & -50.26288 & -34.60857 & -10.67973 & 4.99129 & 27.25994 & 32.42654 & 51.15655 & 53.29083 & 58.69711 & 68.06081\\
\hline
\textbf{$\theta_{pred}^{out}$ (deg)} & -50.27410 & -34.60671 & -10.66992 & 5.00272 & 27.26967 & 32.43521 & 51.15915 & 53.29249 & 58.69611 & 68.05380\\
\hline
\textbf{Absolute error (deg)} & 0.01122 & 0.00186 & 0.00981 & 0.01143 & 0.00973 & 0.00867 & 0.00260 & 0.00166 & 0.00100 & 0.00701\\
\hline
\end{tabular}
}

\resizebox{\textwidth}{!}{%
\begin{tabular}{|c|c c c c c c c c c c|}
\hline
\textbf{Precision points} & \textbf{11} & \textbf{12} & \textbf{13} & \textbf{14} & \textbf{15} & \textbf{16} & \textbf{17} & \textbf{18} & \textbf{19} & \textbf{20}\\
\hline
\textbf{$\theta^{in}$ (deg)} & 39.40456 & 40.64574 & 19.22760 & -5.80374 & -13.88130 & -27.31268 & -32.21906 & -37.27008 & -38.30872 & -39.55164\\
\hline
\textbf{$\theta^{out}$ (deg)} & 77.08304 & 77.55672 & -3.16209 & -43.57679 & -54.40259 & -68.90553 & -72.89030 & -76.07582 & -76.59444 & -77.14375\\
\hline
\textbf{$\theta_{pred}^{out}$ (deg)} & 77.06395 & 77.53592 & -3.17347 & -43.58229 & -54.40377 & -68.89783 & -72.87867 & -76.05929 & -76.57673 & -77.12447\\
\hline
\textbf{Absolute error (deg)} & 0.01909 & 0.02080 & 0.01138 & 0.00550 & 0.00118 & 0.00770 & 0.01163 & 0.01653 & 0.01771 & 0.01928 \\
\hline
\end{tabular}
}
\caption{Precision points, output angles, and error for triple rocker 00$^{+}$ mechanism}
\label{table:Single_example_20_2}
\end{table}

\section{Multi-type Synthesis Using Absolute Precision Points}
\label{sec:Multi-type Synthesis Results}

This section follows a similar structure to \cref{sec:Single-type Synthesis Results}, presenting examples of synthesizing diverse four-bar linkage types simultaneously, referred to as `multi-type synthesis' in this paper.
The objective is to design mechanisms that pass through five or twenty target precision points, while proposing various four-bar linkage types all at once.
As discussed in \cref{sec:Single-type Synthesis Results}, multi-type synthesis is achieved by integrating predictions from all 16 expert models, and thus employs the same simulation metric $S_{simul}$ as organized in \cref{table:Simulation Error Table}.

The examples focus on presenting the top three distinct linkage types that yield the lowest average absolute error across all given target precision points.
While single-type synthesis showcases results that are fully optimized for a single mechanism type, multi-type synthesis combines outputs for multiple types, which requires balancing the performance across all types.
As a result, some precision points may exhibit larger errors, as optimizing the fit for one linkage type can compromise the accuracy for others.

Note that the target precision points in the examples are relatively dense to facilitate the derivation of various types of mechanisms.
While technically more than three linkage types could be generated, in such cases, the precision points would need to be excessively dense for four distinct types to share these points, making the synthesis process effectively meaningless.

\subsection{Five Precision Point Synthesis}
\label{subsec:Five Precision Point Case_multi}

\begin{figure}[htbp]
\centering
\begin{subfigure}[b]{0.45\textwidth}
    \centering
    \includegraphics[width=\textwidth]{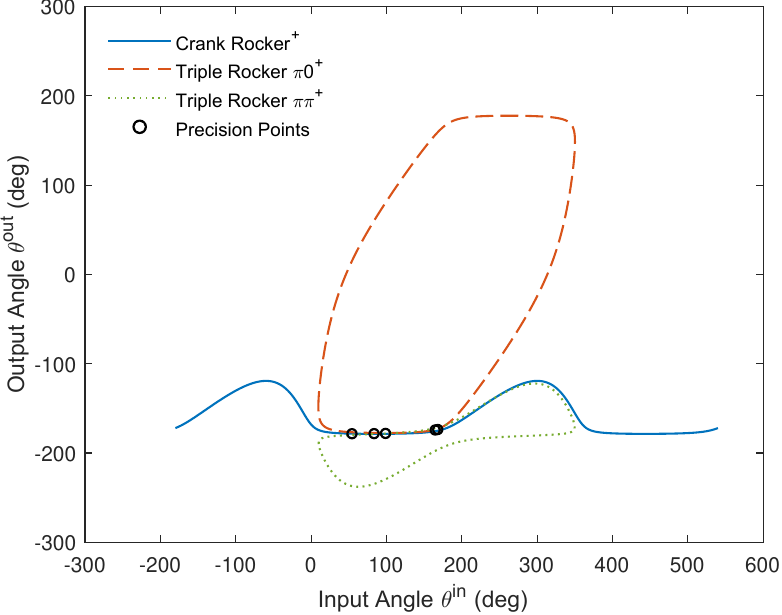}
    \caption*{(a)}
\end{subfigure}
\hfill
\begin{subfigure}[b]{0.45\textwidth}
    \centering
    \includegraphics[width=\textwidth]{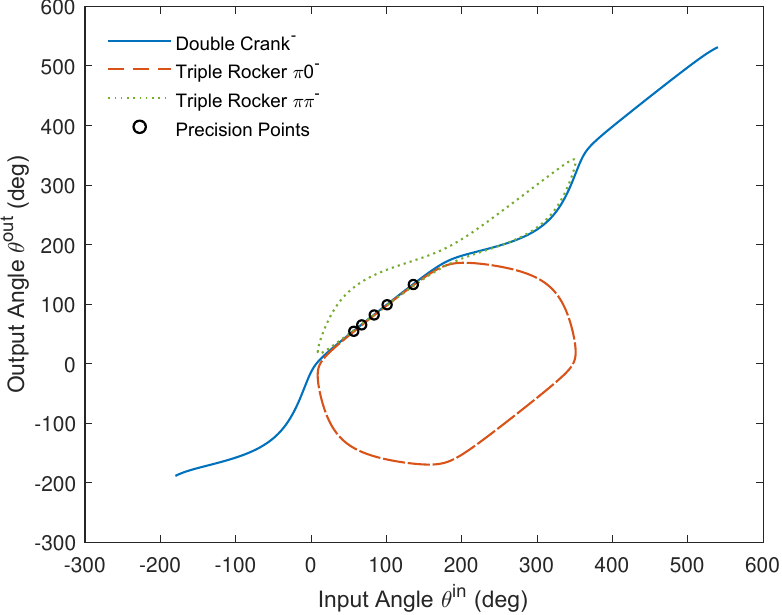}
    \caption*{(b)}
\end{subfigure}
\caption{Displacement plots for five-precision-point synthesis: (a) Case 1 (b) Case 2}
\label{figure:Multi_example_5}
\end{figure}

For the two sets of five arbitrarily given target precision points listed in \cref{table:Multi_example_5_1} and \cref{table:Multi_example_5_2}, three distinct mechanisms are synthesized for each case.
Detailed results, including target precision points, predicted mechanism dimensions, corresponding output angles $(\theta_{pred}^{out})$, and associated absolute errors are also presented in \cref{table:Multi_example_5_1} and \cref{table:Multi_example_5_2}.
The corresponding displacement plots are shown in \cref{figure:Multi_example_5}, confirming that the predicted mechanisms successfully achieve the five target precision points.

\begin{table}[htbp]
\centering
\resizebox{\textwidth}{!}{%
\begin{tabular}{|>{\centering\arraybackslash}p{6.0cm}|c c c c c|}
\hline
\textbf{Precision points} & \textbf{1} & \textbf{2} & \textbf{3} & \textbf{4} & \textbf{5}\\
\hline
\textbf{$\theta^{in}$ (deg)} & 54.32978 & 83.70364 & 98.85742 & 164.75668 & 168.16536\\	
\hline
\textbf{$\theta^{out}$ (deg)} & -178.41287 & -178.31120 & -178.09927 & -174.23944 & -173.69795\\
\hline
\hline
\textbf{Mechanism1 (Crank Rocker$^{+}$)} & \multicolumn{5}{|c|}{$\vec{r}_{pred}$=[3.15884, 1.55958, 1.63900, 3.16858]}\\
\hline
\textbf{$\theta_{pred}^{out}$ (deg)} & -178.35594 & -178.57250 & -178.52442 & -175.63802 & -175.07470\\ 
\hline
\textbf{Absolute error (deg)} & 0.05693 & 0.26130 & 0.42515 & 1.39858 & 1.37675\\
\hline
\hline
\textbf{Mechanism2 (Triple Rocker $\pi$0$^{+}$)} & \multicolumn{5}{|c|}{$\vec{r}_{pred}$=[2.32400, 3.25751, 3.35560, 2.31250]}\\
\hline
\textbf{$\theta_{pred}^{out}$ (deg)} & -176.77231 & -177.51909 & -177.59205 & -173.79573 & -172.92083\\ 
\hline
\textbf{Absolute error (deg)} & 1.64056 & 0.79211 & 0.50722 & 0.44371 & 0.77712\\
\hline
\hline
\textbf{Mechanism3 (Triple Rocker $\pi\pi$$^{+}$)} & \multicolumn{5}{|c|}{$\vec{r}_{pred}$=[3.07914, 1.47759, 1.56266, 3.20781]}\\
\hline
\textbf{$\theta_{pred}^{out}$ (deg)} & -179.84904 & -178.80622 & -178.22320 & -172.17048 & -171.42667\\ 
\hline	
\textbf{Absolute error (deg)} & 1.43617 & 0.49502 & 0.12393 & 2.06896 & 2.27128\\
\hline
\end{tabular}%
}
\caption{Five precision points, output angles, and errors for derived mechanisms (Case1)}
\label{table:Multi_example_5_1}
\end{table}

\begin{table}[htbp]
\centering
\resizebox{\textwidth}{!}{%
\begin{tabular}{|>{\centering\arraybackslash}p{6.0cm}|c c c c c|}
\hline
\textbf{Precision points} & \textbf{1} & \textbf{2} & \textbf{3} & \textbf{4} & \textbf{5}\\
\hline
\textbf{$\theta^{in}$ (deg)} & 56.72772 & 67.27548 & 83.81854 & 100.96771 & 135.72491\\	
\hline
\textbf{$\theta^{out}$ (deg)} & 54.39265 & 65.14343 & 81.82163 & 98.93291 & 132.92754\\
\hline
\hline
\textbf{Mechanism1 (Double Crank$^{-}$)} & \multicolumn{5}{|c|}{$\vec{r}_{pred}$=[2.04794, 3.41266, 2.13080, 3.42700]}\\
\hline
\textbf{$\theta_{pred}^{out}$ (deg)} & 55.22200 & 65.86844 & 82.44932 & 99.51707 & 133.59079\\ 
\hline
\textbf{Absolute error (deg)} & 0.82935 & 0.72501 & 0.62769 & 0.58416 & 0.66325\\
\hline
\hline	
\textbf{Mechanism2 (Triple Rocker $\pi$0$^{-}$)} & \multicolumn{5}{|c|}{$\vec{r}_{pred}$=[3.31579, 2.33531, 3.42436, 2.35532]}\\
\hline
\textbf{$\theta_{pred}^{out}$ (deg)} & 53.85046 & 64.59136 & 81.20748 & 98.20038 & 131.67087\\ 
\hline
\textbf{Absolute error (deg)} & 0.54219 & 0.55207 & 0.61415 & 0.73253 & 1.25667\\
\hline
\hline
\textbf{Mechanism3 (Triple Rocker $\pi\pi$$^{-}$)} & \multicolumn{5}{|c|}{$\vec{r}_{pred}$=[2.02657, 3.49554, 2.16628, 3.69254]}\\
\hline	
\textbf{$\theta_{pred}^{out}$ (deg)} & 56.23209 & 66.29762 & 82.08579 & 98.35749 & 130.43577\\ 
\hline
\textbf{Absolute error (deg)} & 1.83944 & 1.15419 & 0.26416 & 0.57542 & 2.49177\\
\hline
\end{tabular}%
}
\caption{Five precision points, output angles, and errors for derived mechanisms (Case2)}
\label{table:Multi_example_5_2}
\end{table}

\subsection{Twenty Precision Point Synthesis}
\label{subsec:20 Precision Point Case_multi}

\begin{figure}[htbp]
\centering
\begin{subfigure}[b]{0.45\textwidth}
    \centering
    \includegraphics[width=\textwidth]{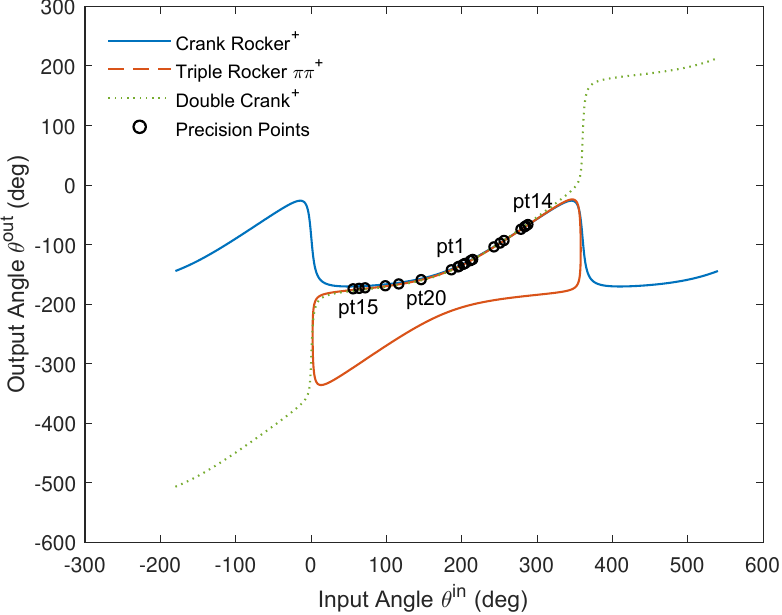}
    \caption*{(a)}
\end{subfigure}
\hfill
\begin{subfigure}[b]{0.45\textwidth}
    \centering
    \includegraphics[width=\textwidth]{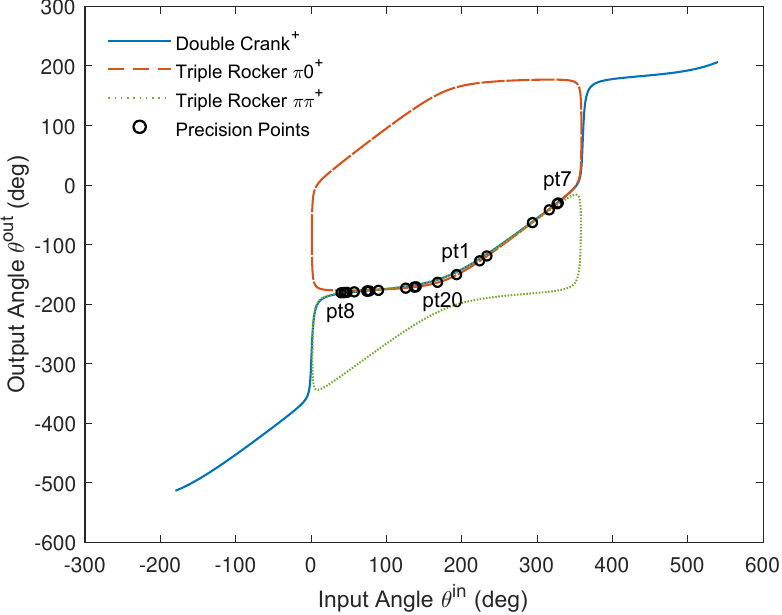}
    \caption*{(b)}
\end{subfigure}
\caption{Displacement plots for twenty-precision-point synthesis: (a) Case 1 (b) Case 2}
\label{figure:Multi_example_20}
\end{figure}

For the twenty precision points given in \cref{table:Multi_example_20_1} and \cref{table:Multi_example_20_2}, results are detailed for three distinct mechanisms, including the twenty given precision points, predicted mechanism dimensions, corresponding output angles $(\theta_{pred}^{out})$, and the corresponding absolute errors.
\cref{figure:Multi_example_20} illustrates the displacement plot, demonstrating the actual output angles of three predicted mechanisms align with target precision points.
Following the same convention as in the single-type synthesis, not all precision points are labeled, though their sequence is accurately arranged.

\begin{table}[htbp]
\centering
\resizebox{\textwidth}{!}{%
\begin{tabular}{|c|c c c c c c c c c c|}
\hline
\textbf{Precision points} & \textbf{1} & \textbf{2} & \textbf{3} & \textbf{4} & \textbf{5} & \textbf{6} & \textbf{7} & \textbf{8} & \textbf{9} & \textbf{10}\\
\hline
\textbf{$\theta^{in}$ (deg)} & -173.79239 & -165.02017 & -163.41753 & -157.98761 & -155.54230 & -148.44261 & -145.49904 & -117.25439 & -109.23280 & -103.86393 \\
\hline
\textbf{$\theta^{out}$ (deg)} & -142.37500 & -137.45702 & -136.50873 & -133.18625 & -131.63652 & -126.96003 & -124.94827 & -103.89111 & -97.46468 & -93.08407 \\
\hline
\hline
\textbf{Mechanism1 (Crank Rocker$^{+}$)} & \multicolumn{10}{|c|}{$\vec{r}_{pred}$=[2.23379, 2.08568, 2.66223, 2.64128]}\\
\hline
\textbf{$\theta_{pred}^{out}$ (deg)} & -141.25642 & -136.45660 & -135.53092 & -132.28726 & -130.77407 & -126.20690 & -124.24174 & -103.66037 & -97.37855 & -93.09797 \\
\hline
\textbf{Absolute error (deg)} & 1.11858 & 1.00042 & 0.97781 & 0.89899 & 0.86245 & 0.75313 & 0.70653 & 0.23074 & 0.08613 & 0.01390 \\
\hline
\hline
\textbf{Mechanism2 (Triple Rocker $\pi\pi$$^{+}$)} & \multicolumn{10}{|c|}{$\vec{r}_{pred}$=[2.18883, 2.16382, 2.60934, 2.69117]}\\
\hline
\textbf{$\theta_{pred}^{out}$ (deg)} & -142.60882 & -137.68480 & -136.73649 & -133.41633 & -131.86881 & -127.20255 & -125.19663 & -104.23792 & -97.85314 & -93.50414 \\
\hline
\textbf{Absolute error (deg)} & 0.23382 & 0.22778 & 0.22776 & 0.23008 & 0.23229 & 0.24252 & 0.24836 & 0.34681 & 0.38846 & 0.42007 \\
\hline
\hline
\textbf{Mechanism3 (Double Crank$^{+}$)} & \multicolumn{10}{|c|}{$\vec{r}_{pred}$=[2.14940, 2.21940, 2.60827, 2.64139]}\\
\hline
\textbf{$\theta_{pred}^{out}$ (deg)} & -143.28777 & -138.23552 & -137.26095 & -133.84565 & -132.25231 & -127.44331 & -125.37424 & -103.70562 & -97.08394 & -92.56563  \\
\hline
\textbf{Absolute error (deg)} & 0.91277 & 0.77850 & 0.75222 & 0.65940 & 0.61579 & 0.48328 & 0.42597 & 0.18549 & 0.38074 & 0.51844    \\
\hline
\end{tabular}
}

\resizebox{\textwidth}{!}{%
\begin{tabular}{|c|c c c c c c c c c c|}
\hline
\textbf{Precision points} & \textbf{11} & \textbf{12} & \textbf{13} & \textbf{14} & \textbf{15} & \textbf{16} & \textbf{17} & \textbf{18} & \textbf{19} & \textbf{20}\\
\hline
\textbf{$\theta^{in}$ (deg)} & -81.43034 & -76.44228 & -75.12250 & -72.21104 & 55.84741 & 63.72791 & 71.69461 & 98.57292 & 116.45685 & 146.12770 \\
\hline
\textbf{$\theta^{out}$ (deg)} & -74.27651 & -70.01773 & -68.88793 & -66.39201 & -174.31091 & -173.52501 & -172.66403 & -169.15228 & -166.09077 & -158.90378 \\
\hline
\hline
\textbf{Mechanism1 (Crank Rocker$^{+}$)} & \multicolumn{10}{|c|}{$\vec{r}_{pred}$=[2.23379, 2.08568, 2.66223, 2.64128]}\\
\hline
\textbf{$\theta_{pred}^{out}$ (deg)} & -74.75529 & -70.61619 & -69.51952 & -67.09915 & -170.29673 & -170.00951 & -169.54082 & -166.87650 & -164.15267 & -157.34068 \\
\hline
\textbf{Absolute error (deg)} & 0.47878 & 0.59846 & 0.63159 & 0.70714 & 4.01418 & 3.5155 & 3.12321 & 2.27578 & 1.93810 & 1.5631     \\
\hline
\hline
\textbf{Mechanism2 (Triple Rocker $\pi\pi$$^{+}$)} & \multicolumn{10}{|c|}{$\vec{r}_{pred}$=[2.18883, 2.16382, 2.60934, 2.69117]}\\
\hline
\textbf{$\theta_{pred}^{out}$ (deg)} & -74.87071 & -70.66317 & -69.54796 & -67.08592 & -176.18011 & -175.12266 & -174.04303 & -170.02886 & -166.74521 & -159.30017 \\
\hline
\textbf{Absolute error (deg)} & 0.59420 & 0.64544 & 0.66003 & 0.69391 & 1.8692 & 1.59765 & 1.37900 & 0.87658 & 0.65444 & 0.39639   \\
\hline
\hline
\textbf{Mechanism3 (Double Crank$^{+}$)} & \multicolumn{10}{|c|}{$\vec{r}_{pred}$=[2.14940, 2.21940, 2.60827, 2.64139]}\\
\hline
\textbf{$\theta_{pred}^{out}$ (deg)} & -73.09012 & -68.65236 & -67.47256 & -64.86186 & -177.60699 & -176.41930 & -175.24378 & -171.05995 & -167.73498 & -160.25102 \\
\hline
\textbf{Absolute error (deg)} & 1.18639 & 1.36537 & 1.41537 & 1.53015 & 3.29608 & 2.89429 & 2.57975 & 1.90767 & 1.64421 & 1.34724 \\
\hline
\end{tabular}
}
\caption{Twenty precision points, output angles, and errors for derived mechanisms (Case1)}
\label{table:Multi_example_20_1}
\end{table}

\begin{table}[htbp]
\centering
\resizebox{\textwidth}{!}{%
\begin{tabular}{|c|c c c c c c c c c c|}
\hline
\textbf{Precision points} & \textbf{1} & \textbf{2} & \textbf{3} & \textbf{4} & \textbf{5} & \textbf{6} & \textbf{7} & \textbf{8} & \textbf{9} & \textbf{10}\\
\hline
\textbf{$\theta^{in}$ (deg)} & -166.87431 & -136.18394 & -126.58101 & -66.12660 & -43.79280 & -33.43674 & -31.99048 & 39.94711 & 43.74075 & 44.65279 \\
\hline
\textbf{$\theta^{out}$ (deg)} & -150.48767 & -127.32546 & -119.08328 & -63.13500 & -41.62692 & -31.52516 & -30.10445 & 178.97906 & 179.37546 & 179.46550 \\
\hline
\hline
\textbf{Mechanism1 (Double Crank$^{+}$)} & \multicolumn{10}{|c|}{$\vec{r}_{pred}$=[2.22404, 2.30393, 2.51560, 2.55950]}\\
\hline
\textbf{$\theta_{pred}^{out}$ (deg)} & -145.99656 & -123.89157 & -116.00270 & -61.53479 & -40.23368 & -30.10954 & -28.67459 & 177.88835 & 178.52341 & 178.66697 \\
\hline
\textbf{Absolute error (deg)} & 4.49111 & 3.43389 & 3.08058 & 1.60021 & 1.39324 & 1.41562 & 1.42986 & 1.09071 & 0.85205 & 0.79853 \\
\hline
\hline
\textbf{Mechanism2 (Triple Rocker $\pi$0$^{+}$)} & \multicolumn{10}{|c|}{$\vec{r}_{pred}$=[2.31670, 2.31556, 2.50653, 2.46171]}\\
\hline
\textbf{$\theta_{pred}^{out}$ (deg)} & -151.08975 & -127.68601 & -119.35095 & -62.78350 & -40.88133 & -30.46158 & -28.98216 & -177.01095 & -177.01353 & -177.01004 \\
\hline
\textbf{Absolute error (deg)} & 0.60208 & 0.36055 & 0.26767 & 0.35150 & 0.74559 & 1.06358 & 1.12229 & 4.00999 & 3.61101 & 3.52446 \\
\hline
\hline
\textbf{Mechanism3 (Triple Rocker $\pi\pi$$^{+}$)} & \multicolumn{10}{|c|}{$\vec{r}_{pred}$=[2.25604, 2.28688, 2.51262, 2.59097]}\\
\hline
\textbf{$\theta_{pred}^{out}$ (deg)} & -146.01976 & -124.20750 & -116.43135 & -63.09291 & -42.80063 & -33.56095 & -32.29303 & 178.46324 & 179.04961 & 179.18268 \\
\hline
\textbf{Absolute error (deg)} & 4.46791 & 3.11796 & 2.65193 & 0.04209 & 1.17371 & 2.03579 & 2.18858 & 0.51582 & 0.32585 & 0.28282 \\
\hline
\end{tabular}
}

\resizebox{\textwidth}{!}{%
\begin{tabular}{|c|c c c c c c c c c c|}
\hline
\textbf{Precision points} & \textbf{11} & \textbf{12} & \textbf{13} & \textbf{14} & \textbf{15} & \textbf{16} & \textbf{17} & \textbf{18} & \textbf{19} & \textbf{20}\\
\hline
\textbf{$\theta^{in}$ (deg)} & 47.65556 & 57.40577 & 74.05494 & 74.89967 & 77.40353 & 89.50002 & 125.68843 & 136.89967 & 139.42667 & 167.89816 \\
\hline
\textbf{$\theta^{out}$ (deg)} & 179.75021 & -179.41331 & -178.12282 & -178.05833 & -177.86662 & -176.91200 & -173.18813 & -171.46895 & -171.01848 & -163.35780 \\
\hline
\hline
\textbf{Mechanism1 (Double Crank$^{+}$)} & \multicolumn{10}{|c|}{$\vec{r}_{pred}$=[2.22404, 2.30393, 2.51560, 2.55950]}\\
\hline
\textbf{$\theta_{pred}^{out}$ (deg)} & 179.11916 & -179.56878 & -177.59536 & -177.49838 & -177.21094 & -175.79957 & -170.60176 & -168.36412 & -167.79399 & -158.94113 \\
\hline
\textbf{Absolute error (deg)} & 0.63105 & 0.15547 & 0.52746 & 0.55995 & 0.65568 & 1.11243 & 2.58637 & 3.10483 & 3.22449 & 4.41667 \\
\hline
\hline
\textbf{Mechanism2 (Triple Rocker $\pi$0$^{+}$)} & \multicolumn{10}{|c|}{$\vec{r}_{pred}$=[2.31670, 2.31556, 2.50653, 2.46171]}\\
\hline
\textbf{$\theta_{pred}^{out}$ (deg)} & -176.98879 & -176.83504 & -176.34564 & -176.31412 & -176.21697 & -175.66508 & -172.85737 & -171.38230 & -170.98528 & -163.84121 \\
\hline
\textbf{Absolute error (deg)} & 3.26100 & 2.57827 & 1.77718 & 1.74421 & 1.64965 & 1.24692 & 0.33076 & 0.08665 & 0.03320 & 0.48341 \\
\hline
\hline
\textbf{Mechanism3 (Triple Rocker $\pi\pi$$^{+}$)} & \multicolumn{10}{|c|}{$\vec{r}_{pred}$=[2.25604, 2.28688, 2.51262, 2.59097]}\\
\hline
\textbf{$\theta_{pred}^{out}$ (deg)} & 179.60315 & -179.16491 & -177.27892 & -177.18530 & -176.90738 & -175.53396 & -170.40321 & -168.18069 & -167.61411 & -158.82479 \\
\hline
\textbf{Absolute error (deg)} & 0.14706 & 0.24840 & 0.84390 & 0.87303 & 0.95924 & 1.37804 & 2.78492 & 3.28826 & 3.40437 & 4.53301 \\
\hline
\end{tabular}
}
\caption{Twenty precision points, output angles, and errors for derived mechanisms (Case2)}
\label{table:Multi_example_20_2}
\end{table}

\section{Multi-type Synthesis Using Relative Precision Points}
\label{sec:Synthesis Using Relative Precision Points}

As discussed in \cref{subsubsec:Precision Points}, this section presents synthesis examples based on relative precision points.
Although relative and absolute precision points may appear equivalent from a dataset generation standpoint, they introduce fundamentally different challenges during model training.

Relative precision points do not specify an initial position for the input and the output links, meaning no ground truth configuration exists.
This absence complicates both supervision and evaluation in data-driven synthesis: without a fixed reference frame, absolute output angles cannot be computed, and relative outputs also remain undefined.
Consequently, any loss function or evaluation metric relying on a known initial configuration becomes unusable, rendering direct training or evaluation with relative precision points infeasible.

To address this issue, an additional processing procedure is incorporated into the proposed framework.
Rather than retraining a separate model, the synthesis problem is reformulated by expanding each sequence of $n$ relative precision points into 100 variants through random assignment of initial configurations.
These variants are then converted into absolute precision points and processed using the same synthesis methodology described in \cref{sec:Data-Driven Method for Dimensional Synthesis}, enabling direct reuse of the trained synthesis model without modification.
Thus, although no absolute precision points are explicitly provided, a single relative-precision-point problem is effectively expanded into 100 absolute-precision-point problems.
Each variant is processed by 16 expert models, resulting in a total of 1,600 predictions per synthesis task, which are subsequently ranked according to the simulation metric $S_{simul}$.
The entire process takes each synthesis task in 0.27 seconds, allowing a throughput of 3.7–3.8 tasks per second.
It is important to note that the actual numerical values of the initial position are not critical, the internal consistency of the resulting motion is essential.

The synthesis results for relative precision points are summarized in \cref{table:Simulation Error Table_rel}, which reports the simulation metrics for all 16 four-bar linkage types across two distinct assembly configurations.

\begin{table}[htbp]
\centering
\begin{tabular}{|l||c|l||c|}
\hline
\multicolumn{1}{|c||}{\textbf{Mechanism Type}} & \multicolumn{1}{c|}{\textbf{Simulation Metric}} & \multicolumn{1}{|c||}{\textbf{Mechanism Type}} & \multicolumn{1}{c|}{\textbf{Simulation Metric}} \\
\hline
Crank Rocker$^{+}$ & 0.0001 & Triple Rocker 00$^{+}$ & 0.0005\\
Crank Rocker$^{-}$ & 0.0001 & Triple Rocker 00$^{-}$ & 0.0005\\
Rocker Crank$^{+}$ & 0.0213 & Triple Rocker 0$\pi$$^{+}$ & 0.0015\\
Rocker Crank$^{-}$ & 0.0199 & Triple Rocker 0$\pi$$^{-}$ & 0.0014\\
Double Crank$^{+}$ & 0.0001 & Triple Rocker $\pi$0$^{+}$ & 0.0009\\
Double Crank$^{-}$ & 0.0001 & Triple Rocker $\pi$0$^{-}$ & 0.0010\\
Double Rocker$^{+}$ & 0.0002 & Triple Rocker $\pi\pi$$^{+}$ & 0.0006\\
Double Rocker$^{-}$ & 0.0002 & Triple Rocker $\pi\pi$$^{-}$ & 0.0004\\
\hline
\end{tabular}
\caption{Simulation metric for 16 types of four-bar mechanisms using relative precision points}
\label{table:Simulation Error Table_rel}
\end{table}

Since multi-type synthesis based on the Mixture of Experts (MoE) framework inherently aggregates the outputs of individual single-type experts, only multi-type synthesis results are presented in the following sections.

\subsection{Seven and Twenty Precision Point Synthesis}
\label{subsec:Seven Precision Point Synthesis}

This section investigates two cases: one using seven relative target precision points, which is the maximum number of points that can be specified as discussed in \cref{subsubsec:Precision Points}, and another using twenty points to demonstrate scalability.
To avoid visual clutter caused by varying initial positions and mechanism types, the results are summarized in \cref{table:Rel_example_4} and \cref{table:Rel_example_19}, while plots are omitted for clarity.
In these tables, $\theta_{precision}^{in}$ and $\theta_{precision}^{out}$ denote the absolute angular positions computed from both the initial position and the relative precision points. 
Notably, the first absolute point in each table represents the initial angles of the input and the output links.
Consistent with \cref{sec:Multi-type Synthesis Results}, the focus is placed on presenting the top three distinct linkage types.

\begin{table}[htbp]
\centering
\resizebox{\textwidth}{!}{%
\begin{tabular}{|c|c c c c c c c|}
\hline
\textbf{Precision points} & &\textbf{2} & \textbf{3} & \textbf{4} & \textbf{5} & \textbf{6} & \textbf{7}\\
\hline
\textbf{$\Delta\theta^{in}$ (deg)} & & -8.70850 & -16.84412 & -31.50208 & -41.79584 & -42.91702 & -70.94623\\	
\hline
\textbf{$\Delta\theta^{out}$ (deg)} & & -4.03400 & -7.75671 & -14.25116 & -18.50137 & -18.94011 & -26.36378\\
\hline
\textbf{Mechanism1 (Crank Rocker$^{+}$)} & \multicolumn{7}{|c|}{$\vec{r}_{pred}$=[1.62822, 2.97689, 0.27608, 2.22574]}\\
\hline
\textbf{Absolute points} & \textbf{1} & \textbf{2} & \textbf{3} & \textbf{4} & \textbf{5} & \textbf{6} & \textbf{7}\\
\hline
\textbf{$\theta_{precision}^{in}$ (deg)} & -134.00734 & -142.71584 & -150.85146 & -165.50942 & -175.80318 & -176.92436 & -204.95357\\ 
\hline
\textbf{$\theta_{precision}^{out}$ (deg)} & -98.63482 & -102.66882 & -106.39153 & -112.88598 & -117.13619 & -117.57493 & -124.99861\\ 
\hline
\textbf{$\theta_{pred}^{out}$ (deg)} & -98.77166 & -103.03572 & -106.82072 & -113.08823 & -117.01885 & -117.42195 & -125.78762\\ 
\hline
\textbf{Absolute error (deg)} & 0.13684 & 0.36690 & 0.42919 & 0.20225 & 0.11734 & 0.15298 & 0.78901\\
\hline
\hline
\textbf{Mechanism2 (Triple Rocker $\pi\pi$$^{-}$)} & \multicolumn{7}{|c|}{$\vec{r}_{pred}$=[1.48288, 3.29028, 0.30164, 1.80585]}\\
\hline
\textbf{Absolute points} & \textbf{1} & \textbf{2} & \textbf{3} & \textbf{4} & \textbf{5} & \textbf{6} & \textbf{7}\\
\hline
\textbf{$\theta_{precision}^{in}$ (deg)} & 164.36043 & 155.65193 & 147.51631 & 132.85836 & 122.56459 & 121.44341 & 93.41420\\ 
\hline
\textbf{$\theta_{precision}^{out}$ (deg)} & 162.62244 & 158.58844 & 154.86573 & 148.37128 & 144.12107 & 143.68233 & 136.25866\\ 
\hline
\textbf{$\theta_{pred}^{out}$ (deg)} & 162.80528 & 158.80290 & 154.96113 & 148.08121 & 143.54473 & 143.07710 & 134.93157\\ 
\hline
\textbf{Absolute error (deg)} & 0.18284 & 0.21446 & 0.09540 & 0.29007 & 0.57634 & 0.60523 & 1.32709\\
\hline
\hline
\textbf{Mechanism3 (Crank Rocker$^{-}$)} & \multicolumn{7}{|c|}{$\vec{r}_{pred}$=[0.08859, 1.65526, 0.92647, 2.54331]}\\
\hline
\textbf{Absolute points} & \textbf{1} & \textbf{2} & \textbf{3} & \textbf{4} & \textbf{5} & \textbf{6} & \textbf{7}\\
\hline
\textbf{$\theta_{precision}^{in}$ (deg)} & 137.47038 & 128.76188 & 120.62626 & 105.96831 & 95.67454 & 94.55336 & 66.52415\\ 
\hline
\textbf{$\theta_{precision}^{out}$ (deg)} & 120.82997 & 116.79597 & 113.07326 & 106.57881 & 102.32860 & 101.88986 & 94.46619\\ 
\hline
\textbf{$\theta_{pred}^{out}$ (deg)} & 120.96378 & 117.23479 & 113.71210 & 107.43086 & 103.22478 & 102.78396 & 93.78826\\ 
\hline	
\textbf{Absolute error (deg)} & 0.13381 & 0.43882 & 0.63884 & 0.85205 & 0.89618 & 0.89410 & 0.67793\\
\hline
\end{tabular}%
}
\caption{Seven relative precision points, output angles, and errors for derived mechanisms}
\label{table:Rel_example_4}
\end{table}

\begin{table}[htbp]
\centering
\resizebox{\textwidth}{!}{%
\begin{tabular}{|c|c c c c c c c c c c|}
\hline
\textbf{Precision points} & & \textbf{2} & \textbf{3} & \textbf{4} & \textbf{5} & \textbf{6} & \textbf{7} & \textbf{8} & \textbf{9} & \textbf{10}\\
\hline
\textbf{$\Delta\theta^{in}$ (deg)} & &0.26543 & 0.43786 & 0.75792 & 1.03045 & 1.16040 & 1.66876 & 1.68249 & 1.79373 & 1.60960 \\
\hline
\textbf{$\Delta\theta^{out}$ (deg)} & &0.14445 & 0.18037 & 0.16507 & 0.07826 & 0.01199 & -0.48734 & -0.51100 & -1.34406 & -1.65047 \\
\hline
\hline
\textbf{Mechanism1 (Double Rocker$^{-}$)} & \multicolumn{10}{|c|}{$\vec{r}_{pred}$=[3.32644, 2.12753, -0.00903, 2.12930]}\\
\hline
\textbf{Absolute points} & \textbf{1} & \textbf{2} & \textbf{3} & \textbf{4} & \textbf{5} & \textbf{6} & \textbf{7} & \textbf{8} & \textbf{9} & \textbf{10}\\
\hline
\textbf{$\theta_{precision}^{in}$ (deg)} & 38.30043 & 38.56586 & 38.73830 & 39.05835 & 39.33088 & 39.46083 & 39.96920 & 39.98293 & 40.09416 & 39.91003\\ 
\hline
\textbf{$\theta_{precision}^{out}$ (deg)} & 142.34464 & 142.48910 & 142.52501 & 142.50971 & 142.42290 & 142.35663 & 141.85730 & 141.83364 & 141.00058 & 140.69417\\ 
\hline
\textbf{$\theta_{pred}^{out}$ (deg)} & 141.48382 & 141.66058 & 141.60380 & 141.31533 & 141.25159 & 141.22063 & 141.09603 & 141.09259 & 141.06456 & 141.11081 \\
\hline
\textbf{Absolute error (deg)} & -0.86082 & -0.82851 & -0.92121 & -1.19438 & -1.17131 & -1.13600 & -0.76128 & -0.74105 & 0.06398 & 0.41664 \\
\hline
\hline
\textbf{Mechanism2 (Double Rocker$^{+}$)} & \multicolumn{10}{|c|}{$\vec{r}_{pred}$=[2.62652, 2.46152, 0.03116, 1.72621]}\\
\hline
\textbf{Absolute points} & \textbf{1} & \textbf{2} & \textbf{3} & \textbf{4} & \textbf{5} & \textbf{6} & \textbf{7} & \textbf{8} & \textbf{9} & \textbf{10}\\
\hline
\textbf{$\theta_{precision}^{in}$ (deg)} & 38.93471 & 39.20013 & 39.37257 & 39.69263 & 39.96515 & 40.09510 & 40.60347 & 40.61720 & 40.72843 & 40.54431\\ 
\hline
\textbf{$\theta_{precision}^{out}$ (deg)} & 115.95103 & 116.09549 & 116.13140 & 116.11610 & 116.02929 & 115.96302 & 115.46369 & 115.44003 & 114.60697 & 114.30056\\ 
\hline
\textbf{$\theta_{pred}^{out}$ (deg)} & 115.39705 & 115.75672 & 115.88679 & 115.97639 & 115.88220 & 115.74552 & 115.29448 & 115.29873 & 115.33293 & 115.27604 \\
\hline
\textbf{Absolute error (deg)} & -0.55398 & -0.33877 & -0.24462 & -0.13971 & -0.14709 & -0.21750 & -0.16922 & -0.14130 & 0.72595 & 0.97548 \\
\hline
\hline
\textbf{Mechanism3 (Crank Rocker$^{-}$)} & \multicolumn{10}{|c|}{$\vec{r}_{pred}$=[1.79867, 0.01728, 2.84374, 1.84381]}\\
\hline
\textbf{Absolute points} & \textbf{1} & \textbf{2} & \textbf{3} & \textbf{4} & \textbf{5} & \textbf{6} & \textbf{7} & \textbf{8} & \textbf{9} & \textbf{10}\\
\hline
\textbf{$\theta_{precision}^{in}$ (deg)} & 86.53655 & 86.80198 & 86.97441 & 87.29447 & 87.56700 & 87.69695 & 88.20531 & 88.21904 & 88.33028 & 88.14615\\ 
\hline
\textbf{$\theta_{precision}^{out}$ (deg)} & 77.78938 & 77.93383 & 77.96975 & 77.95445 & 77.86764 & 77.80137 & 77.30204 & 77.27838 & 76.44532 & 76.13891\\ 
\hline
\textbf{$\theta_{pred}^{out}$ (deg)} & 76.76749 & 76.77048 & 76.77244 & 76.77608 & 76.77920 & 76.78068 & 76.78653 & 76.78669 & 76.78797 & 76.78585 \\
\hline
\textbf{Absolute error (deg)} & -1.02189 & -1.16335 & -1.19732 & -1.17837 & -1.08845 & -1.02070 & -0.51552 & -0.49169 & 0.34265 & 0.64693 \\
\hline
\end{tabular}
}

\resizebox{\textwidth}{!}{%
\begin{tabular}{|c|c c c c c c c c c c|}
\hline
\textbf{Precision points} & \textbf{11} & \textbf{12} & \textbf{13} & \textbf{14} & \textbf{15} & \textbf{16} & \textbf{17} & \textbf{18} & \textbf{19} & \textbf{20} \\
\hline
\textbf{$\Delta\theta^{in}$ (deg)} & 1.23633 & 0.97550 & 0.77390 & 0.65920 & 0.53656 & 0.42612 & 0.30923 & 0.14206 & -0.12497 & -0.22095\\
\hline
\textbf{$\Delta\theta^{out}$ (deg)} & -1.86461 & -1.89433 & -1.87024 & -1.83939 & -1.79236 & -1.73679 & -1.66283 & -1.52440 & -1.17096 & -0.93953\\
\hline
\hline
\textbf{Mechanism1 (Double Rocker$^{-}$)} & \multicolumn{10}{|c|}{$\vec{r}_{pred}$=[3.32644, 2.12753, -0.00903, 2.12930]}\\
\hline
\textbf{Absolute points} & \textbf{11} & \textbf{12} & \textbf{13} & \textbf{14} & \textbf{15} & \textbf{16} & \textbf{17} & \textbf{18} & \textbf{19} & \textbf{20}\\
\hline
\textbf{$\theta_{precision}^{in}$ (deg)} & 39.53676 & 39.27593 & 39.07433 & 38.95963 & 38.83699 & 38.72655 & 38.60966 & 38.44249 & 38.17546 & 38.07948\\ 
\hline
\textbf{$\theta_{precision}^{out}$ (deg)} & 140.48003 & 140.45031 & 140.47440 & 140.50525 & 140.55228 & 140.60785 & 140.68181 & 140.82024 & 141.17368 & 141.40511\\ 
\hline
\textbf{$\theta_{pred}^{out}$ (deg)} & 141.20237 & 141.26458 & 141.31163 & 141.33801 & 141.49532 & 141.61176 & 141.65753 & 141.61660 & 141.51033 & 141.53045 \\
\hline
\textbf{Absolute error (deg)} & 0.72234 & 0.81427 & 0.83723 & 0.83275 & 0.94304 & 1.00390 & 0.97572 & 0.79636 & 0.33665 & 0.12533\\
\hline
\hline
\textbf{Mechanism2 (Double Rocker$^{+}$)} & \multicolumn{10}{|c|}{$\vec{r}_{pred}$=[2.62652, 2.46152, 0.03116, 1.72621]}\\
\hline
\textbf{Absolute points} & \textbf{11} & \textbf{12} & \textbf{13} & \textbf{14} & \textbf{15} & \textbf{16} & \textbf{17} & \textbf{18} & \textbf{19} & \textbf{20}\\
\hline
\textbf{$\theta_{precision}^{in}$ (deg)} & 40.17104 & 39.91021 & 39.70861 & 39.59391 & 39.47127 & 39.36083 & 39.24394 & 39.07677 & 38.80974 & 38.71376\\ 
\hline
\textbf{$\theta_{precision}^{out}$ (deg)} & 114.08642 & 114.05670 & 114.08079 & 114.11164 & 114.15867 & 114.21424 & 114.28820 & 114.42663 & 114.78007 & 115.01150\\ 
\hline
\textbf{$\theta_{pred}^{out}$ (deg)} & 115.60370 & 115.91766 & 115.97575 & 115.96916 & 115.93470 & 115.87986 & 115.79581 & 115.62026 & 115.07899 & 114.62934\\
\hline
\textbf{Absolute error (deg)} & 1.51728 & 1.86097 & 1.89496 & 1.85752 & 1.77603 & 1.66562 & 1.50760 & 1.19363 & 0.29892 & -0.38216\\
\hline
\hline
\textbf{Mechanism3 (Crank Rocker$^{-}$)} & \multicolumn{10}{|c|}{$\vec{r}_{pred}$=[1.79867, 0.01728, 2.84374, 1.84381]}\\
\hline
\textbf{Absolute points} & \textbf{11} & \textbf{12} & \textbf{13} & \textbf{14} & \textbf{15} & \textbf{16} & \textbf{17} & \textbf{18} & \textbf{19} & \textbf{20}\\
\hline
\textbf{$\theta_{precision}^{in}$ (deg)} & 87.77288 & 87.51205 & 87.31045 & 87.19575 & 87.07311 & 86.96267 & 86.84578 & 86.67861 & 86.41158 & 86.31560\\ 
\hline
\textbf{$\theta_{precision}^{out}$ (deg)} & 75.92477 & 75.89505 & 75.91914 & 75.94999 & 75.99702 & 76.05259 & 76.12655 & 76.26498 & 76.61842 & 76.84985\\ 
\hline
\textbf{$\theta_{pred}^{out}$ (deg)} & 76.78155 & 76.77856 & 76.77626 & 76.77495 & 76.77356 & 76.77230 & 76.77098 & 76.76909 & 76.76608 & 76.76500 \\
\hline
\textbf{Absolute error (deg)} & 0.85678 & 0.88351 & 0.85712 & 0.82496 & 0.77654 & 0.71971 & 0.64443 & 0.50411 & 0.14766 & -0.08485 \\
\hline
\end{tabular}
}
\caption{Twenty relative precision points, output angles, and errors for derived mechanisms}
\label{table:Rel_example_19}
\end{table}

\section{Conclusions}
\label{sec:Conclusions}

This research proposes a novel data-driven approach for the dimensional synthesis of four-bar function generators, leveraging machine learning to overcome the limitations of conventional analytical and optimization-based methods.
By utilizing a type-specified dataset, the proposed method trains 16 specialized LSTM expert models, each corresponding to a specific four-bar linkage type.
These expert models are integrated through a Mixture of Experts (MoE) framework, forming a unified system capable of capturing the unique motion characteristics of each linkage type.
Regardless of the number of precision points, this architecture enables both single-type and multi-type synthesis within a unified system, eliminating the need for additional training or manual adjustments.
The synthesis results demonstrate high accuracy for both absolute and relative precision point specifications, effectively generating multiple high-quality, defect-free solutions for specified linkage types, and enabling the simultaneous synthesis of diverse mechanism types within a reasonable time frame.
In addition to its standalone utility, this framework also serves as a valuable source of initial solutions for traditional optimization-based approaches, allowing for more refined tuning where needed.

Overall, this study establishes a robust and user-friendly methodology that offers an accessible alternative to conventional synthesis approaches, even for users without prior expertise in kinematics or machine learning.
Collectively, this work not only demonstrates the feasibility of data-driven dimensional synthesis but also lays the groundwork for more intelligent and adaptable design tools in mechanical system development.

\section{Discussion}
\label{sec:Discussion}

This study was conducted using the model architecture and training environment described in \cref{subsubsec:Implementation Details}, which adopts a relatively simple model configuration selected to perform the synthesis task efficiently and with minimal computational overhead.
The results confirm the framework's practical effectiveness, while also indicating that more sophisticated model architectures could yield even lower simulation metrics—assuming a feasible solution exists—by producing more accurate linkage configurations.
In other words, the proposed method offers a solid foundation, but the performance has not yet reached the theoretical limits, highlighting opportunities for future enhancements in model design, training strategies, and dataset construction.

From a broader perspective, this framework can be extended to other mechanism synthesis problems, such as path generation and motion generation, by leveraging well-established kinematic formulations that support the generation of appropriate training datasets.
In path generation, the model would learn mappings between the tracer point positions and the mechanism dimensions $\vec{r}$.
For motion generation, an additional parameter representing the orientation of the floating link can be included as part of the output.
Furthermore, the proposed framework is extendable to mechanisms with a greater number of links—such as six-bar or eight-bar linkages—provided that the corresponding kinematic analysis equations are available for dataset construction.
Additionally, the framework could be adapted to incorporate transmission angle constraints, enabling the design of more efficient and realistic mechanisms tailored to specific performance criteria.
While these extensions may require minor tuning of the model configuration, the proposed framework shows strong potential for generalization across a wide range of mechanism types and tasks.

\bibliographystyle{ieeetr}
\bibliography{References}

\end{document}